%% file: A-main-IEEE.tex
\documentclass[10pt,final]{IEEEtran}

\usepackage{mathtools}
\usepackage{graphicx}
\graphicspath{{figs/}}
\usepackage[utf8]{inputenc} % allow utf-8 input
\usepackage[T1]{fontenc}    % use 8-bit T1 fonts
\usepackage[dvipsnames]{xcolor}         % colors
\usepackage[noadjust]{cite}
\usepackage[bookmarks=true,breaklinks=true,colorlinks,linkcolor=RedViolet,citecolor=blue,urlcolor=BlueViolet]{hyperref}
\usepackage{url}            % simple URL typesetting
\usepackage{booktabs}       % professional-quality tables
\usepackage{amsmath,amssymb,amsfonts,amsthm,stmaryrd}
\usepackage{nicefrac}       % compact symbols for 1/2, etc.
\usepackage{microtype}      % microtypography
\usepackage{authblk}
\usepackage[ruled]{algorithm2e}
\usepackage[noend]{algorithmic} 
\usepackage{algorithmic}

\usepackage{subcaption}
\usepackage{dblfloatfix}
\usepackage{dsfont}
\usepackage[misc]{ifsym}

\usepackage{siunitx}
\input{C-macro.tex}

\title{Heterogeneous Graph Prompt Learning via Adaptive Weight Pruning }

\author{
  Chu-Yuan~Wei,~
  Shun-Yao~Liu,~
  Sheng-Da~Zhuo,~
  Chang-Dong~Wang, \IEEEmembership{Senior Member,~IEEE,}~
  and~
  Shu-Qiang~Huang~and Mohsen~Guizani,~\IEEEmembership{Fellow,~IEEE}
   \thanks{
    C.-Y. We is College of Electrical and Information Engineering, Beijing University of Civil Engineering and Architecture, Beijing, China.
    E-mail:  weichuyuan@bucea.edu.cn
  } 
    \thanks{
    S.-Y. Liu is College of Electrical and Information Engineering, Beijing University of Civil Engineering and Architecture, Beijing, China.
    E-mail:  2108550022064@stu.bucea.edu.cn
  } 
  \thanks{
    S.-D. Zhuo is College of Cyber Security of Jinan University, Jinan University, Guangzhou,  China. 
    E-mail: Zhuosd96@gmail.com
  } 
  \thanks{
    C.-D. Wang is with School of Computer, Sun Yat-sen University, Guangzhou, China. 
    E-mail: changdongwang@hotmail.com
  } 
  \thanks{
  S.-Q. Huang is with College of Cyber Security of Jinan University, Jinan University, 
  and also with Guangdong Key Laboratory of Data Security and Privacy Preserving, Guangzhou, China. 
  E-mail:hsq@jnu.edu.cn
  }
 \thanks{M. Guizani is with the Machine Learning Department, Mohamed Bin Zayed University of Artificial Intelligence (MBZUAI), Abu Dhabi, UAE. E-mail:\href{mailto:mguizani@ieee.org}{mguizani@ieee.org}.}

  \thanks{
    Corresponding Authors: Shu-Qiang Huang (hsq@jnu.edu.cn)
  }
}

\begin{document}
\pagenumbering{gobble} % Show/not page numbers
\maketitle
\thispagestyle{plain} 

\begin{abstract}
Graph Neural Networks (GNNs) have achieved remarkable success in various graph-based tasks (\emph{e.g.}, node classification or link prediction). 
Despite their triumphs, GNNs still face challenges such as long training and inference times, difficulty in capturing complex relationships, and insufficient feature extraction. 
To tackle these issues, graph pre-training and graph prompt methods have garnered increasing attention for their ability to leverage large-scale datasets for initial learning and task-specific adaptation, offering potential improvements in GNN performance.
% These have led to increased attention on graph pre-training and graph prompt methods for their ability to leverage large-scale datasets for initial learning and adaptation to specific tasks. 
However, previous research has overlooked the potential of graph prompts in optimizing models, as well as the impact of both positive and negative graph prompts on model stability and efficiency.
To bridge this gap, we propose a novel framework combining graph prompts with weight pruning, called \alg, which aims to enhance the performance and efficiency of graph prompts by using fewer of them. 
We evaluate the importance of graph prompts using an importance assessment function to determine positive and negative weights at different granularities. 
Through hierarchically structured pruning, we eliminate negative prompt labels, resulting in more parameter-efficient and competitively performing prompts. 
Extensive experiments on three benchmark datasets demonstrate the superiority of \alg, leading to a significant reduction in parameters in node classification tasks.
%achieving an average parameter reduction of 55.04\% in node classification tasks.
\end{abstract}

\input{B-1-Introduction.tex}

\input{B-3-Preliminaries.tex}
\input{B-2-Related.tex}

\input{B-4-Model.tex}

\input{B-5-Experiments-A.tex}

\input{B-6-Conclusion.tex}

\section*{Acknowledgement}
This work was supported by the national key research and development program “Industrial Software” key special project “Collaborative Optimization and Dynamic Game Decision Making of Parts Supply Chain Product Service Life Cycle Process” under Grant 2022YFB3305602, 
the Humanities and Social Sciences Planning Fund of the Ministry of Education under Grant 22YJAZH110,
National Natural Science Foundation of China (No.62272198, 62276277), 
Guangdong Key Laboratory of Data Security and Privacy Preserving under Grant  2023B1212060036, 
Guangdong-Hong Kong Joint Laboratory for Data Security and Privacy Preserving under Grant  2023B1212120007, 
Guangdong Basic and Applied Basic Research Foundation under Grant 2024A1515010121, 
and 
the Special Funds for the Cultivation of Guangdong College Students’ Scientific and Technological Innovation (Climbing Program Special Funds) under Grant pdjh2025ak028.
% and the Outstanding Innovative Talents Cultivation Funded Programs for Doctoral Students of Jinan University under Grant 2023CXB022. 

Any opinions, findings, and conclusions expressed in this publication are those of the authors and do not necessarily reflect the views of the funding agencies.

\bibliographystyle{IEEEtran}
\bibliography{D-Ref-List}

\end{document}

%% file: C-macro.tex
\usepackage{mathtools}
\usepackage{graphicx}
\graphicspath{{figs/}}
\usepackage[utf8]{inputenc} % allow utf-8 input
\usepackage[T1]{fontenc}    % use 8-bit T1 fonts
\usepackage[dvipsnames]{xcolor}         % colors
\usepackage[noadjust]{cite}
\usepackage[bookmarks=true,breaklinks=true,colorlinks,linkcolor=RedViolet,citecolor=blue,urlcolor=BlueViolet]{hyperref}
\usepackage{url}            % simple URL typesetting
\usepackage{booktabs,multirow}       % professional-quality tables
\usepackage{amsmath}
\usepackage{amssymb}
\usepackage{amsfonts}
\usepackage{stmaryrd}
\usepackage{nicefrac}       % compact symbols for 1/2, etc.
\usepackage{microtype}      % microtypography
\usepackage{authblk}
\usepackage{subcaption}
\usepackage{dblfloatfix}
\usepackage{dsfont}
\usepackage[misc]{ifsym} % Add the Letter sign
\usepackage[inline,shortlabels]{enumitem} % Enumerate inline
\usepackage{diagbox}
\usepackage{siunitx}
\renewcommand{\eqref}[1]{Eq.\,(\ref{#1})}

\newcommand{\yA}[1]{\hat{y}_{\text{obs}}}
\newcommand{\yR}[1]{\hat{y}_{\text{rec}}}

\newcommand{\romup}[1]{\uppercase\expandafter{\romannumeral #1\relax}}
\newcommand{\romlo}[1]{\lowercase\expandafter{\romannumeral #1\relax}}

\newcommand{\rom}[1]{\uppercase\expandafter{\romannumeral #1\relax}}
\newcommand{\romsm}[1]{\lowercase\expandafter{\romannumeral #1\relax}}

\newcommand{\alg}{{GPAWP}} % OS2LMD
\usepackage[inline,shortlabels]{enumitem} % Enumerate inline
\renewcommand{\eqref}[1]{Eq.\,(\ref{#1})}

%% file: B-1-Introduction.tex
% ===================
% # I. Introduction #
% ===================
\section{Introduction}
\label{sec: intro}
Graph data has garnered significant attention for its ability to represent complex relationships and networks, particularly in data analysis scenarios requiring high levels of interconnectedness, such as social network analysis~\cite{li2023survey}, recommendation systems~\cite{xia2022multi}, bioinformatics~\cite{reau2023deeprank}, and cybersecurity~\cite{he2022illuminati}. 
The intrinsic characteristics of graph data, including its relational structure, flexibility, and dynamic nature, enable it to represent and analyze complex inter-entity relationships effectively.
The ability to model these intricate relationships, however, also presents unique challenges when working with limited labeled data.

Traditional graph neural networks (GNNs)~\cite{wu2020comprehensive} typically require a large amount of labeled data for effective training. In scenarios with limited labeled data~\cite{chen2020smoothing,hu2023cost,he2023label}, GNNs trained from scratch may struggle to perform well. 
However, pre-training and fine-tuning strategies~\cite{cheng2023wiener,jiang2021pre,jin2021node} can significantly improve this issue. 
The pre-training phase, often involving unsupervised or self-supervised learning, provides a solid foundation for feature extraction~\cite{chen2020generative}. 
 Pre-training offers a robust base by learning rich feature representations from large-scale datasets~\cite{dong2019unified}, thereby enhancing the model's generalization capabilities. 
The subsequent fine-tuning phase further optimizes the model on this foundation, allowing it to better adapt to specific task requirements.

\begin{figure}[!t]
    \centering
    \includegraphics[width=\columnwidth]{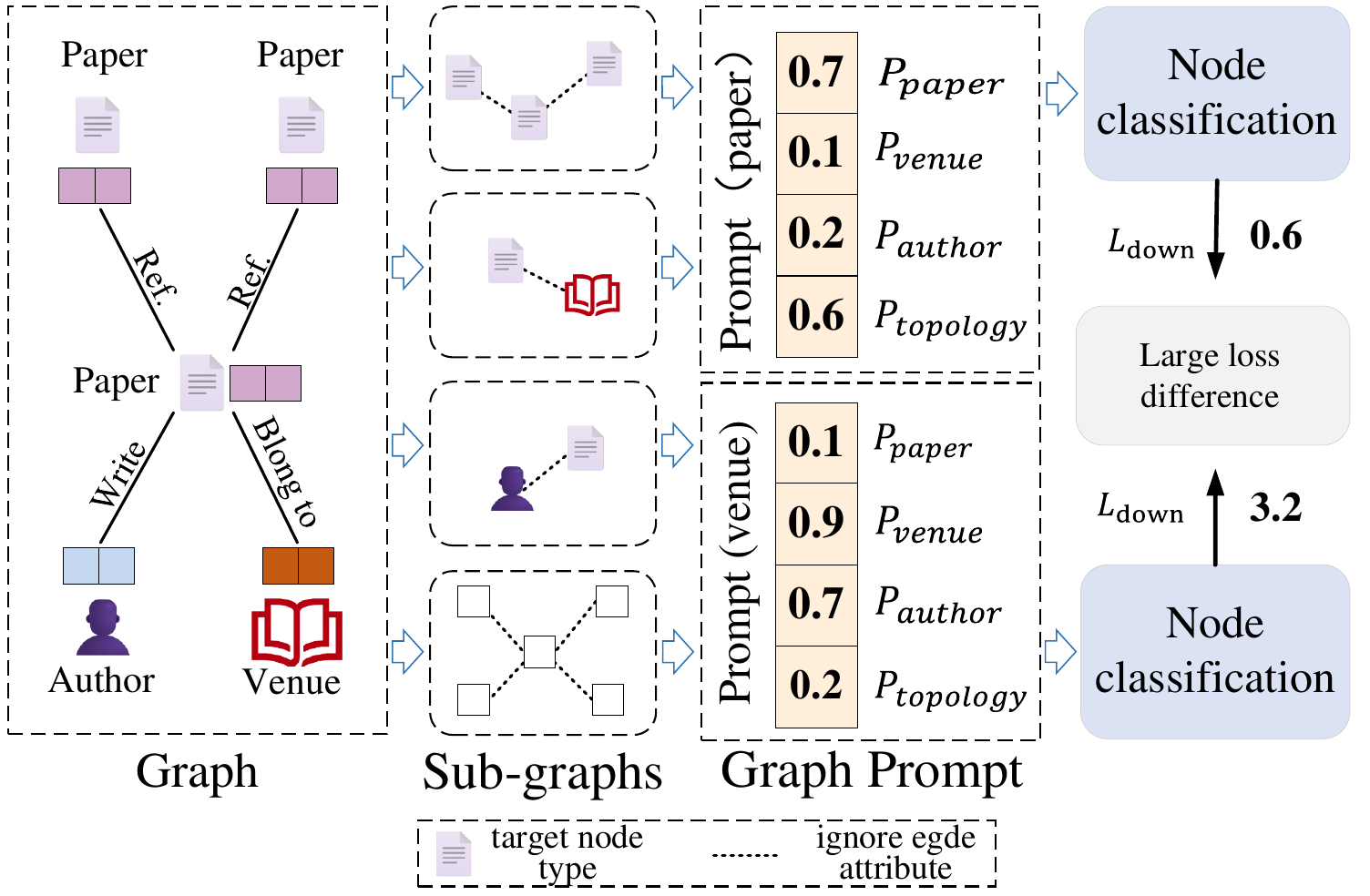}
    \caption{Illustrates the significant impact of different graph prompts on node classification loss, highlighting the critical importance of prompt design in graph prompt learning. 
    (Ref. is the short name for Reference.)}
    \label{fig:motivation}
\end{figure}

Inspired by the success of pre-trained large language models (LLM) in the field of Natural Language Processing (NLP) ~\cite{brown2020language,liu2021p}, the concept of Prompt Learning is gradually gaining traction in the domain of graph data. 
Extending prompt learning to graph data~\cite{sun2022gppt,liu2023graphprompt,ge2023domain} aims to enhance the performance of GNNs on specific tasks without necessitating extensive retraining or comprehensive model adjustments. 
Graph prompt learning involves designing task-related graph structure prompts and integrating them into the original graph~\cite{yu2024hgprompt,ma2023hetgpt,lv2025graphprompter}, allowing for task-specific model adjustment or fine-tuning, with performance evaluation to verify the effectiveness of the prompts. Although graph prompt learning can exhibit impressive performance in certain scenarios~\cite{fang2022prompt,tan2023virtual}, fine-tuning parameters for large-scale graph data~\cite{mh2024lvm,li2023scigraphqa} can be inefficient due to the memory overhead proportional to the number of trainable parameters required to store gradients and optimizer states. 
% As shown in Fig. (\ref{fig:motivation}), there may be graph prompts with different weights leading to significantly different results in the final node classification task.
Moreover, current research has not thoroughly evaluated the effectiveness of the employed graph prompts, nor has it considered that some graph prompts or parts of graph prompts might not effectively bridge the gap between downstream tasks and pre-training tasks. 
Take the task of node classification in graphs as an example, as shown in Figure~\ref{fig:motivation}.
The figure illustrates how varying graph prompts influence node classification. Assigning different weights to node types like ‘Paper’ and ‘Venue’ results in significant differences in classification loss. Lower weights for key nodes such as ‘Paper’ reduce classification accuracy, highlighting the importance of carefully balancing prompt weights to optimize performance.
% (Fig. \ref{fig:motivation}). 
%\textcolor{blue}{ The "Paper" node represents the target node type for the node classification task. Graph prompts typically incorporate information based on assigned weights, which influence their contribution to the task. If the weight corresponding to the "Paper" node in the prompt is too low, the prompt’s ability to assist in accurately classifying the "Paper" node diminishes. Varying the weights in graph prompts can lead to substantially different classification outcomes. The key scientific question this raises is how to effectively balance and optimize the weight assignments in graph prompts to enhance the accuracy and robustness of node classification, particularly for the target node type.}

To address these issues, we propose a novel graph prompt learning framework, namely \textbf{G}\emph{raph} \textbf{P}\emph{rompt Learning via} \textbf{A}\emph{daptive }\textbf{W}\emph{eight} \textbf{P}\emph{runing} (\alg), 
% \textbf{M}\emph{ore Powerful Graph} \textbf{Prompt} \emph{Learning with Fewer Prompts} (\alg), 
designed to enhance the effectiveness of graph prompts and the efficiency of the graph prompt tuning phase. 
Specifically, we define a graph prompt importance assessment function to evaluate the graph prompts. By distinguishing positive prompts from negative prompts based on the importance scores corresponding to the graph prompts, we then use a pruning strategy to remove the negative prompts, while the retained positive prompts undergo comprehensive graph prompt tuning.
This ensures that fewer graph prompts carry more informative prompts, maximizing the efficiency of graph prompt utilization and making prompts more effective with fewer parameters. 
Additionally, reducing graph prompt parameters decreases the number of parameters required for prompt tuning, thereby ensuring model efficiency.
This paper makes the specific \textbf{contributions}, summarized as follows:
\begin{itemize}[leftmargin=*]
    \item We initiate and investigate the challenging problem of graph prompt effectiveness in large-scale graph data, which can accommodate many real-world downstream tasks but has not been explored in previous work.
    \item We propose a novel design for evaluating and pruning graph prompts, wherein we introduce a graph prompt importance assessment function to handle the varying significance of different graph prompts, and then adjust the graph prompt learning model using positive prompts to better enhance downstream task performance. 
    \item We conduct extensive empirical evaluations on three real-world benchmark datasets to demonstrate the superiority of our proposed graph prompt strategy.
\end{itemize}

In summary, Section \ref{sec: intro} introduces the background of the article and our contributions. 
Section \ref{sec: Challenge} describes the challenges of the research problems and the motivation for the study. 
Section \ref{sec:prelimi} introduces the background of the model used in this article, the problem definition in the pre-training phase, and the definition of the heterogeneous graph template used in \alg. 
Section \ref{sec:related work} provides an overview of the graph pre-training and the graph prompt learning.
Section \ref{sec:model} proposes the method and model design of this paper, including the specific design and implementation of the graph prompt importance evaluation function and graph prompt pruning strategy.
Section \ref{sec:exp} describes the experimental design, dataset selection, and model training and evaluation details. 
Section \ref{sec:conclusion} summarizes the key findings and contributions of the study.

%% file: B-3-Preliminaries.tex
\section{Challenges and Motivation}
\label{sec: Challenge}
Due to the diverse types of graph prompts, many of which types of graph prompts play different roles in downstream tasks. Therefore, there are still challenges in unifying the evaluation and pruning of these prompts.

\par\smallskip\noindent
\textbf{Challenge I - Evaluating the Importance of Graph Prompts.} 
\emph{How should we design important strategies to evaluate the significance of different types of graph prompts? }
Graph prompt learning~\cite{gao2024protein} aims to address the shortage of labeled data and bridge the gap between pre-training tasks and downstream tasks. 
Unlike NLP prompts, the effectiveness of graph prompts cannot be intuitively assessed, and there is no appropriate evaluation method~\cite{ma-etal-2022-xprompt}. 
Existing methods often rely on metrics like accuracy, precision, and recall~\cite{liu2020towards} to indirectly evaluate the effectiveness of graph prompts~\cite{fang2024universal}, rather than using more direct and detailed evaluation strategies. This indirect approach focuses on overall model performance but doesn't clearly show how specific aspects of the graph prompts, such as weight distribution, affect the outcome. One reason for this is the complexity of graph structures, which makes it hard to assess the impact of individual graph prompts directly. Isolating the contribution of a graph prompt to a model's performance requires additional analysis, and there are no established benchmarks for direct evaluation in this area. As a result, this indirect evaluation can lead to problems. It limits the understanding of how graph prompts actually contribute to the task, making it difficult to optimize their design. Additionally, it makes diagnosing issues like poor weight distribution challenging, as the focus is on the overall model output rather than the prompt's specific role. Developing more direct evaluation methods could provide clearer insights into the role of graph prompts and lead to more effective optimization.

In the language domain, methods like soft prompts in NLP~\cite{ma-etal-2022-xprompt} and attention heads in Transformers offer some insights, but graph prompts differ due to their structural and semantic diversity~\cite{michel2019sixteen}. 
Previous methods, focused on single objects, are unsuitable for graph prompts. We propose a graph prompt evaluation method that assesses different types of graph prompts. 
For subgraphs, importance evaluation strategies calculate each part's contribution to tasks like node classification~\cite{shen2020network} or link prediction~\cite{wang2020neighborhood}. For feature prompts, we divide them into finer units to evaluate the impact of individual features, highlighting the most relevant ones and reducing redundancy. 
This approach enhances interpretability and improves model performance by refining the prompt design and isolating key contributions.

\par\smallskip\noindent
\textbf{Challenge II - Pruning Negative Graph Prompts.} 
\emph{After obtaining the importance metrics of prompts, how can we leverage different fine-grained prompts to enhance the overall performance of GNN models? }
Previous research has not focused on evaluating the importance of graph prompts, making it an open question to effectively use this information to improve model performance.
% Inspired by the Lottery Ticket Hypothesis (LTH)~\cite{DBLP:conf/iclr/FrankleC19}, which aims to achieve more efficient model training and inference by identifying and utilizing "lucky subnetworks" within large networks, we introduce the Prompt Pruning operation in \alg. 
% Previous research has not focused on evaluating the importance of graph prompts, making it an open question as to how to effectively use this information to improve model performance.
Inspired by the Lottery Ticket Hypothesis (LTH)~\cite{DBLP:conf/iclr/FrankleC19}, which posits that within large neural networks exist smaller, sparse subnetworks capable of achieving comparable performance when properly trained, we apply this concept to graph neural networks.
we propose a graph prompt pruning mechanism to remove less useful or negative graph prompts. This approach allows the model to eliminate irrelevant or misleading information, thereby focusing on the most impactful prompts. By dynamically pruning negative prompts, we enhance the model's ability to discriminate between key features and noise. Simultaneously, positive prompts are optimized to guide the model in focusing on essential features, improving its ability to learn from relevant samples. This dynamic optimization process not only increases the efficiency of graph prompts but also makes the model more adaptable and task-specific. Ultimately, GPAWP aims to create a framework that evaluates, prunes, and optimizes graph prompts, leading to more efficient graph prompt usage and improved GNN performance across various tasks.

% By combining the positive and negative benefits of different prompts, we optimize the prompt information for GNNs, enhancing the model's adaptability and performance across different tasks.
% First, negative prompt information is used to guide the model in identifying and eliminating unimportant or misleading features. 
% In \alg, we perform pruning operations on negative prompts to enhance the model's discriminative ability. 
% Second, positive prompt information aims to re-guide the model to focus on learning key features from positive samples, enhancing the model's ability to recognize positive samples. 
% Thus, by dynamically pruning and optimizing prompt information, we can further improve the efficiency of graph prompts and the performance of downstream tasks.

\par\smallskip\noindent
\textbf{Motivation.} 
Graph prompt learning has emerged as a key method for extracting useful information from vast graph data, especially when labeled data is scarce. 
However, existing approaches treat all graph prompts as equally important~\cite{yu2024hgprompt,liu2023graphprompt}, using fixed parameters that hinder optimization and limit model performance. 
In reality, the importance of graph prompts varies significantly, and understanding this difference is crucial for improving GNN performance.
The lack of direct evaluation methods for graph prompts leads to inefficient optimization. 
Current approaches rely on indirect metrics, which obscure how individual prompts contribute to tasks. 
A more refined evaluation framework is needed to assess the importance of different types of prompts, whether they involve subgraphs or node features, enabling targeted improvements in prompt design.
Once the graph prompts importance is evaluated, another challenge is how to effectively utilize this information. Inspired by the LTH~\cite{DBLP:conf/iclr/FrankleC19}, we propose adaptive Prompt Pruning to eliminate less useful or negative graph prompts. 
This allows the graph model to focus on the most impactful graph prompts, improving its efficiency and adaptability across tasks.
In summary, the goal of \alg~is to develop a framework that evaluates and optimizes graph prompts, enhancing GNN performance by making graph prompt usage more efficient and task-specific.

\section{Preliminaries}
\label{sec:prelimi}
This section first describes the background of the pre-training GNN method we use and the problem definition of the pre-training task, then describes the definition of the heterogeneous graph templates used in \alg.

\subsection{Graph Pre-training} \label{graph pre-training}
\begin{figure}[!t] 
    \centering
    \includegraphics[width=0.85\columnwidth]{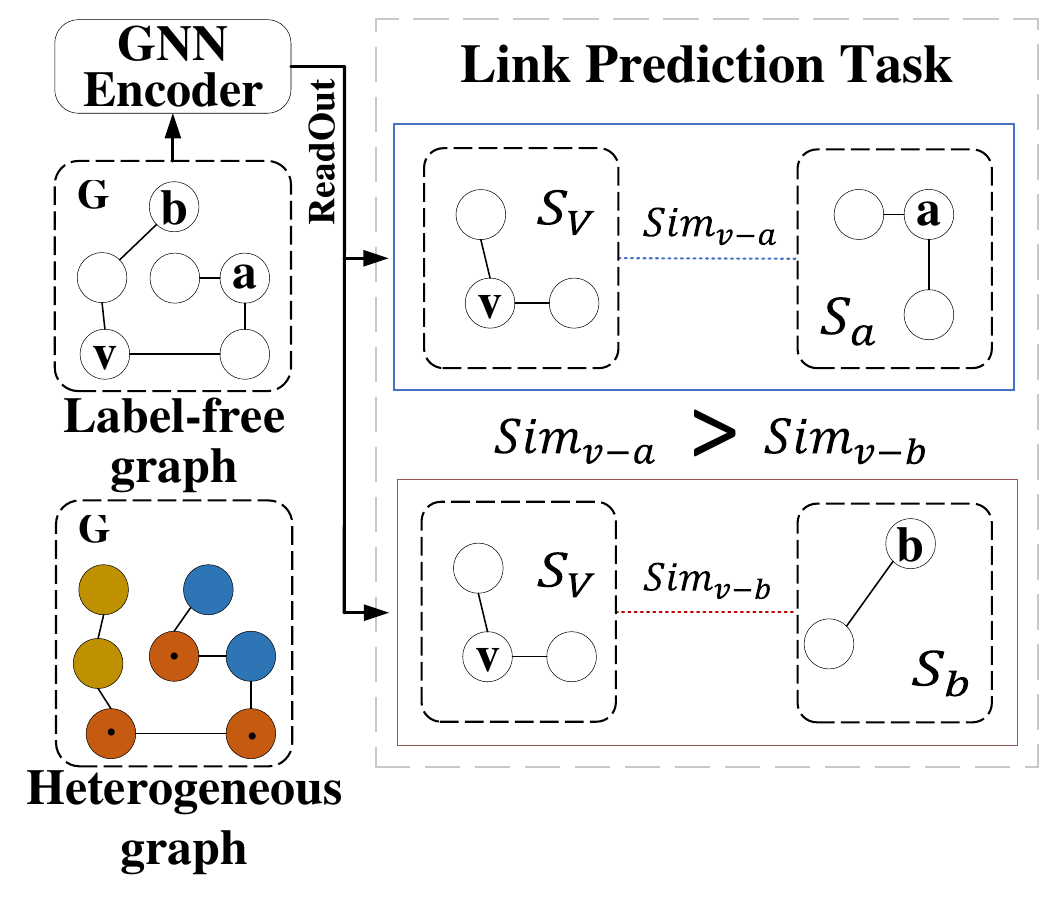}
    \caption{Illustration of GNN-based pre-training method for link prediction. In the label-free graphs, the GNN encoder generates node representations and compares similarities (\emph{e.g.}, $Sim_{v-a}$ vs. $Sim_{v-b}$). In the heterogeneous graph, different colored nodes  
    represent different node types.}
    \label{fig:pretrain}
\end{figure}

% \textbf{Definition 3: Graph Pre-training.} 
Existing graph pre-training methods are divided into node-level, edge-level, graph-level, and multi-task pre-training. 
This paper mainly focuses on edge-level pre-training tasks: \textit{Link prediction} (LP)~\cite{liu2023graphprompt}. 
Given a graph \(G\) and a node triplet \(\left ( v,a,b \right ) \), where\(\left ( v,a \right ) \) is an edge present in \(G\) and \(\left ( v,b \right ) \) is not.
So it exists:
\begin{equation} 
    \label{formula 1}
    sim\left (  s_{v},s_{a}\right ) > sim\left (  s_{v},s_{b}\right )\:
\end{equation}
where \(sim\left ( \cdot,\cdot\right )\) is a similarity function and we use the cosine similarity function. The nodes \(a\) and \(v\) are connected so the similarity between their respective subgraphs is greater than the subgraphs formed by each of \(b\) and \(v\) which are not connected.  During the graph pre-training stage, use Eq. (\ref{formula 1}) to convert the LP task into a similarity comparison task, our goal is to increase the similarity between \(s_{v}\) and \(s_{a}\) while decreasing the similarity between \(s_{v}\) and \(s_{b}\), in Fig. \ref{fig:pretrain} we briefly describe this process. 
Specifically, we sample some triples similar to \(\left ( v, a, b \right)\) from a set of label-free graphs \textit{G}. 
These triples construct an overall training set \(\mathcal{M}_{\text {pre}}\), so the pre-training loss function can be defined as: 
\begin{equation}
    \small
    \mathcal{L}_{\text {pre }}(\Theta)=-\sum_{(v, a, b) \in \mathcal{M}_{\text {pre }}} \ln \frac{\exp \left(\operatorname{sim}\left(\mathbf{s}_{v}, \mathbf{s}_{a}\right) / \tau\right)}{\sum_{u \in\{a, b\}} \exp \left(\operatorname{sim}\left(\mathbf{s}_{v}, \mathbf{s}_{u}\right) / \tau\right)}\:
\end{equation}
where \(\tau\) is a temperature hyperparameter, and \(\Theta\) is the parameters of the model used in the pre-training stage, such as GCN~\cite{kipf2016semi}, GAT~\cite{velickovic2017graph}, \emph{etc}.
The best parameter weight \(\Theta\) saved during pre-training will be used as the initial parameter weight of the model during downstream tasks.
\subsection{ Heterogeneous graph template} \label{Heterogeneous graph template}
% \textbf{Definition 1: Heterogeneous graph. }

% We explain the definition of the heterogeneous graph~\cite{ma2023hetgpt}.
A heterogeneous graph is \( {G} = \left (  {V} ,  {E} \right)  \), where \( {V}\) is the set of nodes and \( {E}\) is the set of edges. There is also a node type mapping function \(\phi\): \(  {V}\to  {A} \) and an edge type mapping function \(\varphi \): \(  {E}\to   {R}\), where \( {A}\) and \( {R}\) denote the node type set and edge type set, respectively, such that \(\left |  {A}  \right | +  \left |   {R} \right | > 2\). The input of the nodes is defined by us as a feature matrix: \(\mathrm {X} \in \mathbb{R}^{\left | V \right | \times d} \).

The heterogeneous graph template is used to process heterogeneous graphs into a unified form~\cite{yu2024hgprompt}. 
In detail, it converts a heterogeneous graph \( {G} = \left (  {V},  {E} \right)  \) into multiple homogeneous subgraphs according to the type of nodes. 
For example, there exists a node type \(i \in A\), we obtain a homogeneous subgraph  \(G^{i}= \left (V^{i},E^{i}\right )  \) of type i from \textit{G}. 
Through this method, a heterogeneous graph \textit{G} can be converted into multiple homogeneous graphs \(\hat{G} = \left \{ G^{i} \mid i \in A \right \} \). 
To preserve the interactions between different types, when \(i = 0\), \(G^{i}\) denotes the complete topology of the graph \textit{G} but does not distinguish between node types. 
Therefore, a heterogeneous graph is divided into \(|\textit{A}|+1\) homogeneous subgraphs through the graph template.

%% file: B-2-Related.tex
\section{Related Work}
\label{sec:related work}
% To address the challenge of limited labeled data in GNNs and to bridge the gap that exists between pre-training tasks and downstream tasks, the pre-training and prompt tuning paradigm, which has flourished in the field of NLP, has been widely adopted in graph representation learning~\cite{sun2023graph}. In order to gain a deeper understanding of the application of this paradigm to the graph domain, we will introduce it in the following two ways:

In this section, we outline two key technologies: \emph{Graph Pre-training} and \emph{Graph Prompt Learning}. 
% Graph Pre-training enhances the model's generalization capabilities on graph data through pre-training; Graph Prompt Learning optimizes prompt mechanisms to tailor the model for specific tasks. 
The combination of these methods effectively boosts the efficiency and performance of graph data processing.

% \par\smallskip\noindent
% \textbf{Graph Pre-training. } 
\subsection{Graph Pre-training}
Graph pre-training~\cite{you2020graph,ju2022commonsense} aims to capture and utilize the information within graph structures through model pre-training to enhance performance on downstream tasks such as node classification~\cite{li2024graph}, graph classification~\cite{wei2021pooling}, or link prediction~\cite{tan2023bring}. 
The core of this approach is leveraging the structural information of graph data to improve the model's generalization capabilities and efficiency. 
The development of graph pre-training has evolved from direct training to incorporating self-supervised learning and then to developing pre-training frameworks similar to BERT~\cite{devlin2018bert} in NLP. 
This method uses self-supervised tasks, such as node prediction and graph reconstruction, which do not require labeled data, to train graph models and learn a generic representation of graph structures. 
Researchers have explored cross-domain transferability~\cite{liu2021learning,chen2022cross} as technology has progressed and gradually applied these techniques to specific problems like molecular design~\cite{sun2022does} and social network~\cite{zhang2022robust} analysis. 

Graph pre-training not only improves model performance on specific tasks, particularly in scenarios with scarce labeled data but also enhances the capability to handle large-scale graph data, demonstrating significant potential in practical applications. 
However, despite these advances, graph pre-training has notable limitations. 
First, it requires extensive pre-training on large datasets to be effective, which is computationally expensive and time-consuming. 
Second, once the pre-training is done, the model often becomes static, lacking adaptability to specific downstream tasks. 
These models may not always generalize well to new or changing data conditions, especially in scenarios with limited labeled data.
While graph pre-training improves general performance, it faces challenges in efficiency, scalability, and adaptability to task-specific data. 
This highlights the need for methods that can dynamically adjust to new tasks, which leads us to the exploration of graph prompt learning.

% Inspired by pre-training models in the NLP domain~\cite{beltagy2019scibert}~\cite{dong2019unified} and the computer vision(CV)~\cite{bao2021beit} domain, graph pre-training methods~\cite{cheng2023wiener}~\cite{jiang2021pre}~\cite{jin2021node} have sprung up. This approach uses off-the-shelf information to encode the inherent graph structure in a self-supervised manner, providing a solid foundation for generalization across multiple downstream tasks. In order to implement specific downstream tasks, it is also necessary to update the pre-trained weights for the task and complete fine-tuning. Existing graph pre-training methods can be categorized into node-level~\cite{hamilton2017inductive}, edge-level~\cite{pan2018adversarially}, graph-level~\cite{you2020graph} and multi-task~\cite{zhang2020graph} pre-training strategies. 

% Due to the differences between the pretraining task and downstream tasks, transferring knowledge from the pretraining task to downstream tasks can often lead to negative transfer~\cite{rosenstein2005transfer}. 
% The impact of negative transfer can lead to a decrease in performance on downstream tasks, potentially resulting in the model performing even worse than random learning on those tasks.

% \par\smallskip\noindent
% \textbf{Graph Prompt Learning. }
\subsection{Graph Prompt Learning}
In the field of NLP, prompt learning is often used as an alternative to fine-tuning and can bridge the gap between the pre-trained task and the downstream task. Thus the paradigm of "pre-training, prompt tuning" has been invoked in the field of graphical representation learning to solve the same problem. Prompts in NLP are usually textual prompts that are combined with the input text to allow for re-customization of the inputs, making the model better adapted to downstream tasks.
Existing methods dealing with homogeneous graphs such as GPPT~\cite{sun2022gppt} and GraphPrompt~\cite{liu2023graphprompt} use self-defined prompt templates to transform both pre-training tasks and downstream tasks into similarity prediction tasks. 
The graph prompt in All in One~\cite{sun2023all} introduces additional subgraphs that can be efficiently learned through tuning. Similarly, PRODIGY~\cite{huang2024prodigy} utilizes a prompt graph consisting of data graphs and a task graph.
ULTRA-DP~\cite{chen2023ultra} employs prompt tuning both during pre-training and downstream tasks, swiftly refining task embeddings during pre-training, and integrating them as a crucial component of the inferred class with the highest similarity in inference results. These methods, however, are limited to homogeneous graphs and rely on predefined prompt templates, which may restrict flexibility in more complex or heterogeneous graphs.

The graph prompting methods for handling heterogeneous graphs are HGPrompt~\cite{yu2024hgprompt} and HetGPT~\cite{ma2023hetgpt}, the former converts heterogeneous graphs into multiple homogeneous graphs by using customized graph templates and then unifies the pre-training task with the downstream task by using the task templates in the Graphprompt, and the latter extends the prompt tokens into a format specific to types, enabling it to address graph prompting within heterogeneous datasets. However, none of these methods evaluate or screen the validity of the graph prompts used.
This absence of prompt evaluation can lead to inefficient prompt designs and suboptimal model performance.
% Similarly, but not identical, methods for graph prompt learning can be divided into two categories due to the fact that they deal with different types of graphs, such as heterogeneous and homogeneous graphs. 
% Existing methods dealing with homogeneous graphs such as GPPT~\cite{sun2022gppt} and GraphPrompt~\cite{liu2023graphprompt} use self-defined prompt templates to transform both pre-training tasks and downstream tasks into similarity prediction tasks. ULTRA-DP~\cite{chen2023ultra} employs prompt tuning both during pre-training and downstream tasks, swiftly refining task embeddings during pre-training, integrating them as a crucial component of the inferred class with the highest similarity in inference results. The graph prompting methods for handling heterogeneous graphs are HGPrompt~\cite{yu2024hgprompt} and HetGPT~\cite{ma2023hetgpt}, the former converts heterogeneous graphs into multiple homogeneous graphs by using customized graph templates, and then unifies the pre-training task with the downstream task by using the task templates in the Graphprompt, and the latter extends the prompt tokens into a format specific to types, enabling it to address graph prompting within heterogeneous datasets. However, none of these methods evaluate or screen the validity of the graph prompts used.\\
Compared to the static nature of pre-trained graph models, graph prompt learning significantly enhances model adaptability and flexibility to new tasks and changing data conditions through dynamic prompt adjustments.
Graph prompt learning offers a solution by allowing dynamic adjustments to specific tasks, but it suffers from a lack of prompt evaluation strategies. This paper aims to address this gap by proposing methods to assess and optimize graph prompts, improving the adaptability and performance of GNN models in various downstream tasks.
% Although graph prompt learning with fewer prompts may not match the performance of comprehensively pre-trained models, it can quickly achieve competitive results on specific tasks through carefully designed prompts, offering a balance between efficiency and effectiveness.

% Compared to the static nature of pre-trained graph models, graph prompt learning significantly enhances model adaptability and flexibility to new tasks and changing data conditions through dynamic prompt adjustments. 
% Although graph prompt learning with fewer prompts may not match the performance of comprehensively pre-trained models, it can quickly achieve competitive results on specific tasks through carefully designed prompts, offering a balance between efficiency and effectiveness.

%% file: B-4-Model.tex
% =============================================
% # IⅤ. Modeling and consistency validations #
% =============================================
\section{Our \alg~Approach}
\label{sec:model}
\label{sec:framework}
\begin{figure*}[!t] 
    \centering
    \includegraphics[width=1\textwidth]{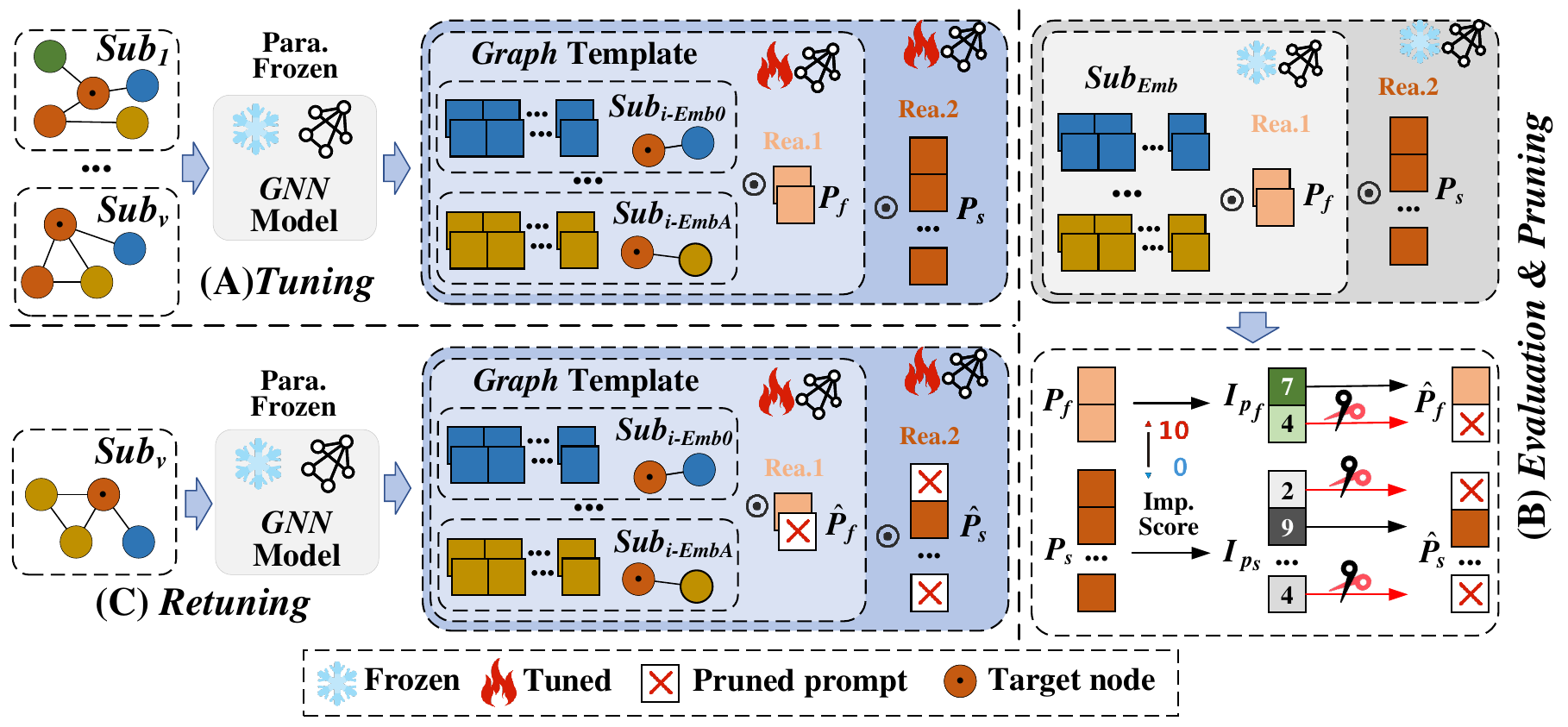}
    \caption{Overall framework of \alg. (A) \emph{Tuning} refers to the tuning of graph prompts on the node task. (B) \emph{Evaluation \& Pruning} is the evaluation of graph prompts and the pruning of negative graph prompts. (C) \emph{Retuning} is the retuning of pruned graph prompts.
    (Sub, Para., and Imp. are short names for Subgraph, Parameter, and Importance).
    }
    \label{fig:model}
\end{figure*}

In this section, we will elaborate on our \alg~framework (Fig. \ref{fig:model}), which is divided into three key modules: \textbf{\textit{Tuning}}, \textit{\textbf{Evaluation \& Pruning}}, and \textbf{\textit{Retuning}}. These modules are designed to address different challenges in graph prompt learning and enhance the overall performance of GNN models. (A) \emph{Tuning}: This module addresses the challenge of feature extraction and embedding refinement, ensuring that the model captures both structural and semantic information. The most effective graph prompt parameters are frozen and retained for further optimization in the next phases. (B) \emph{Evaluation \& Pruning}: In this module, a scoring function is introduced to measure the significance of each graph prompt token, helping to identify which graph prompts contribute most to the task. The pruning operation removes less important graph prompts, ensuring graph prompt efficiency and reducing redundancy. This step supports the subsequent \emph{Retuning} phase by refining the set of graph prompts used for optimization.  (C) \emph{Retuning}: The final module aims to optimize positive graph prompts further, improving the model's performance in the downstream node classification task. This module builds on the results of the \emph{Evaluation \& Pruning} phase, refining the model’s focus and ensuring better performance in node classification.

Each module is interdependent: \emph{Tuning} provides the initial embeddings and graph prompt configurations, \emph{Evaluation \& Pruning}  optimizes the graph prompt set by removing less important graph prompts, and \emph{Retuning} ensures that the model can make the most of the pruned prompts to achieve higher performance. Together, these modules create a systematic approach to refining graph prompts, ensuring their adaptability and effectiveness for the node classification task.

\subsection{Tuning}
% \subsection{Prompt Tuning}
\label{sec:prompt tuning}
 Graph prompt tuning after pre-training has been proven to be effective for both graph classification and node classification tasks. In our graph prompt tuning stage, we conduct a full prompt tuning on the node classification task to obtain the feature prompt tokens $P_f$ and semantic prompt tokens $P_s$~\cite{yu2024hgprompt}. The purpose of the methods used during the graph prompt tuning phase is to formulate the downstream task as a pre-training task ~\cite{tan2023virtual,chen2023ultra}. We employ a task template ~\cite{liu2023graphprompt} to convert NC task into a similarity prediction task akin to the one employed in handling LP task. Given a graph\textit{ G}, a set of node classes \textit{C}, and a labeled node set \(\mathcal{M} = \left \{ \left ( v_{1}, l_{1} \right ), \left (  v_{2}, l_{2} \right ), \ldots \right \} \), where \(\left ( v_{i}, l_{i} \right )\) is the \textit{i}-th node and the class label of this node. For node class \(c\), \(c \in C\), a virtual subgraph is constructed as the node class prototype, and its embedding can be expressed as a vector \(\tilde{\mathbf{s}}_{c}\):
\begin{equation}\label{eq.3} 
    \tilde{\mathbf{s}}_{c}=\frac{1}{k} \sum_{\left(v_{i}, \ell_{i}\right) \in \mathcal{M}, \ell_{i}=c} \mathbf{s}_{v_{i}}\:
\end{equation}

In this way, for a node \(v_j\) without a label, its class label \(l_j\) can be predicted as:
\begin{equation} \label{formula 4}
    l_{j} = arg\:max_{c\in C }\:sim(s_{v_{j} },\tilde{s}_{c})\:
\end{equation}

A node should belong to the class whose prototype subgraph is most similar to the contextual subgraph of the node. Through Eq. (\ref{formula 4}), the NC task is formulated as a similarity prediction task similar to the pre-training task. 

\par\smallskip\noindent
{\bf Feature prompt.}  Feature prompts are often used to reformulate the inputs of a pre-trained model so that the pre-trained model can better handle downstream node classification task. We combine feature prompts with the embeddings of nodes in the input graph to form a new input, so that the downstream task can better use the features of the nodes in the graph through the feature prompts~\cite{yu2024hgprompt}. Specifically, we let $P_f$ denote the vector that can be learned by the downstream task. As shown in Fig. \ref{fig:model}(A), the ReadOut operation performed on some subgraph \textit{S} is as follows:
\begin{equation} \label{eq.5}
 \operatorname{ReadOut}\left(\left\{{P}_f  \odot {h}_{v} \mid v \in V(S)\right\}\right)\:
\end{equation}
where \(\odot\) denotes the element-wise multiplication, \({h}_{v}\) denotes the embedding obtained from the output of node \textit{v} at the last layer of the GNN. Note that the prompt $P_f$ has the same dimension as the node embeddings. The choice of the aggregation scheme for ReadOut is sum pooling.

\par\smallskip\noindent
{\bf Semantic prompt.} We utilized a graph template to partition a heterogeneous graph into homogeneous subgraphs for downstream processing of the heterogeneous graph \textit{S}. The semantic prompt corresponds to the importance of these subgraphs in the current downstream task. Specifically, \(P_s=\left(p_{s}^{{0}}, p_{s}^{ {1 }}, \ldots, p_{s}^{ {|\textit{A}| }}\right)\) is a set of vectors that can be learned, \(p_{s}^{\text {0}}\) corresponds to \(G^{{0}}\) obtained from graph template with complete graph topology information but does not distinguish node types. So the prompt vector has \(|\textit{A}|+1\) dimensions. The aggregation formula of \(P_s\) and each homogeneous subgraph is :
\begin{equation}   \label{6}
\operatorname{ReadOut}\left ( \left \{ \left ( 1\:+\:p_{s}^{\text {i }} \right )  \odot\operatorname{ReadOut}\left( S^{i} \right)\mid S^{i}\in \hat{G}\left (  S\right )   \right \}  \right )\:
\end{equation}

In summary, given a labeled training set \(\mathcal{M} = \left \{ \left ( x_{1} , y_{1} \right ), \left (  x_{2}, y_{2} \right ), \ldots \right \} \), where \(x_i\) is a node, \(y_i\) is the class label of the \(x_i\). The loss function in the prompt tuning stage can be constructed as follows:
\begin{equation}\label{7}
\small
\mathcal{L}_{\text {down }}\left(P_f,P_s\right)=
 -\sum_{\left(x_{i}, y_{i}\right) \in \mathcal{M}_{\text {down }}} \ln \frac{\exp \left(\frac{1}{\tau} \operatorname{sim}\left(\mathbf{s}_{x_{i}}, \tilde{\mathbf{s}}_{y_{i}}\right)\right)}{\sum_{c \in C} \exp \left(\frac{1}{\tau} \operatorname{sim}\left(\mathbf{s}_{x_{i}}, \tilde{\mathbf{s}}_{c}\right)\right)} \:
\end{equation}
Where \(\tilde{\mathbf{s}}_{c}\) is the class prototype embedding generated according to Eq. (\ref{eq.3}), and prompt tuning loss is only parameterized by \(P_f\) and \(P_s\). The output of the prompt tuning phase is the optimal prompts \((P_f, P_s) = arg\:min\mathcal{L}_{\text {down }}(P_f, P_s)\), which will be used for subsequent prompt \emph{Evaluation \& Pruning}.
\subsection{Evaluation \& Pruning}\label{sec:prompt evaluation and pruning}

The goal of graph prompt evaluation and pruning is to remove prompts that negatively impact performance and identify the most effective ones for the current downstream task. This process lays the foundation for further optimization in subsequent stages. As shown in Fig. \ref{fig:model}(B), we first perform the prompt evaluation on the optimal prompts $P_s$ and $P_f$ obtained in the \emph{Tuning} phase, and then prompt pruning is performed based on the evaluation results to retain only those that are most beneficial for the current node classification task.

\par\smallskip\noindent
{\bf Prompt Evaluation.}
\label{sec:4.3.1}
Our approach draws on soft prompt evaluation methods~\cite{ma-etal-2022-xprompt}. The semantic prompts and feature prompts used in this paper play different roles in downstream node classification task, we need to evaluate their importance separately. Since semantic prompts correspond to each type of subgraph, their dimensions correspond to the number of subgraphs obtained from the graph template. Feature prompts correspond to the dimensions of the node feature embeddings. Therefore, the length of semantic prompts is much shorter than that of feature prompts, allowing direct evaluation based on each token. When evaluating $P_s$, we first associate a mask variable \(\lambda_i\) to each $p_{s}^{i}$:
 \begin{equation} \label{eq p2p^}
    \hat{P}_s= \lambda \cdot P_s \:
 \end{equation}
where \(\lambda=\left\{\lambda_{1}, \lambda_{2}, \ldots, \lambda_{n}\right\}, \lambda_{i} \in\{0,1\}  \), and \(\lambda_{i}=0\) indicates that the corresponding  prompt token $p_{s}^{i}$ is pruned. Next, we use the importance evaluation function to determine the importance of each prompt token in $P_s$. The importance function is defined as the expected sensitivity~\cite{michel2019sixteen} of the model to the mask variable \(\lambda\). Specifically, if we need to evaluate the importance score of $p_{s}^{i}$, we need to set \(\lambda_i\) to 0 and all other \(\lambda\) values to 1. This represents the degree of impact on model loss when $p_{s}^{i}$ is pruned, which can be approximated as the importance of this part of the prompt. The importance score $I_{p_{s}^{i}}$ corresponding to each $p_{s}^{i}$ in $P_s$ is defined as:
\begin{equation} \label{ips}
    I_{p_{s}^{i}}=\mathbb{E}_{x \sim \mathcal{M}_{x}}\left|\frac{\partial \mathcal{L}_{\text {down }}(x)}{\partial \lambda_{i}}\right| \:
\end{equation}
where \(\mathcal{M}_{x}\) is the data distribution and \(\mathcal{L}_\text{down}(x)\) is the loss on sample \(x\). A higher $I_{p_{s}^{i}}$ means that pruning away its corresponding prompt token may have a greater impact on the model.

% Therefore, we can easily judge which semantic prompt tokens are effective for the overall model to deal with current downstream task based on the importance function of $P_s$. When the importance score of a prompt token is too small, it means that pruning the prompt will have little impact on the model and can even improve the overall performance of the model. Specifically, if $p_{s}^{i}$ has a low importance score, it means that the information carried by its corresponding subgraph can be ignored for solving downstream tasks. On the contrary, if $p_{s}^{i}$ has a high importance score, then the information carried by the corresponding subgraph needs to be valued because it is very important for solving downstream tasks.
We can easily judge which semantic prompt tokens are effective for the overall model to deal with current downstream task based on the importance function of $P_s$. If a semantic prompt token receives a low importance score, it indicates that the subgraph it represents does not contribute significant information to the task at hand. The model can effectively ignore the relationships or structural patterns within that subgraph without losing performance. Removing such tokens can reduce noise, leading to a cleaner input signal for the model. This pruning step eliminates redundancy and allows the model to focus on more relevant patterns, improving overall performance. Specifically, if $p_{s}^{i}$ has a low importance score, it means that the information carried by its corresponding subgraph can be ignored for solving downstream tasks. Conversely, if a prompt token has a high importance score, it signifies that the information carried by the corresponding subgraph is crucial for solving the downstream task. This may involve essential structural relations, node types, or connectivity patterns that are key to making accurate predictions. Retaining and prioritizing these high-importance tokens ensures that the model captures the most relevant graph features necessary for solving the node classification problem effectively.
If $p_{s}^{i}$ has a high importance score, then the information carried by the corresponding subgraph needs to be valued because it is very important for solving downstream tasks.

When evaluating feature prompts, the approach we use differs from the method applied to semantic prompts. While the evaluation of semantic prompts can be more straightforward due to their structural focus, feature prompts carry far more granular information related to node characteristics. Because these node features are often multidimensional and complex, a simple evaluation method would not capture the detailed influence of each feature on the model’s performance. The need for a fine-grained evaluation of feature prompts arises from the fact that node features typically contain various types of information, such as numerical values, categorical attributes, or even embeddings from external data sources. These features do not contribute equally to the downstream task, and some might have a stronger impact on classification accuracy than others. In summary, fine-grained prompt evaluation for feature prompts is necessary because of the diverse nature of node features and their varying levels of importance for the task. This evaluation ensures that the model leverages only the most relevant and effective features. 
We adapt the block-wise pruning strategy from XPROMPT \cite{ma-etal-2022-xprompt}—a method originally developed for text prompts by partitioning them into contiguous chunks and masking each chunk based on learned importance scores—to evaluate feature prompts in graph neural networks, thereby enabling fine-grained analysis of node feature contributions.
Specifically, we divide the feature prompt vector of $P_f$ into \(t\) blocks in equal proportions, \(P_f = \left \{ P_f^{1},P_f^{2}, \ldots,P_f^{t} \right \} \), we then associate a mask variable \(\eta \) to each \(P_f^{j}\):
\begin{equation}\label{eq pf2pf^}
\hat{P}_f= \eta \cdot P_f \:
\end{equation}
where \(\eta=\left\{\eta_{1}, \eta_{2}, \ldots, \eta_{k}\right\}, \eta_{j} \in\{0,1\}  \), \(\eta_{j}=0\) indicates that the corresponding prompt block is pruned. The formula for the importance score $I_{p_{f}^{j}}$ of \(P_f^{j}\) is: 
\begin{equation} \label{ipf}
I_{p_{f}^{j}}=\mathbb{E}_{x \sim \mathcal{M}_{x}}\left|\frac{\partial \mathcal{L}_{\text {down }}(x)}{\partial \eta_{j}}\right| \:
\end{equation}
we use $I_{p_{f}^{j}}$ to judge which parts of $P_f$ are useful for downstream tasks. The lower (higher) the $I_{p_{f}^{j}}$, the weaker (stronger) the prompting effect of the corresponding prompt block. 

\begin{algorithm}[!t]
    \caption{Prompt Tuning}
    \label{alg:algorithm 1}
    \renewcommand{\algorithmicrequire}{\textbf{Input:}}
    \renewcommand{\algorithmicensure}{\textbf{Output:}}

    \begin{algorithmic}[1]
        \REQUIRE \parbox[t]{.9\linewidth}{A heterogeneous graph \( {G} = \left (  {V},  {E} \right)  \) with multiple node types and edge types, labeled set \(\mathcal{M} = \left \{ \left ( x_{i}, y_{i} \right )\mid j = 1,2 \ldots \right \} \), class set \textit{C}, node type set \textit{A}, pre-trained GNN model \(f_{\Theta_{best}} \) that takes a graph as input and outputs embeddings for the nodes in the graph.}
        \ENSURE   \parbox[t]{.9\linewidth}{feature prompt \(P_f\),  semantic prompt \(P_s\).}

        \STATE \(\text {Initialize graph prompts }P_{f}, P_{s}. \)
        \STATE \( \text{Obtain the embeddings of graph \( {G}\) by \(f_{\Theta_{best}} \) }. \) 

        \WHILE{not convert}
            \FOR{each \(i \in A + 1\)}
                \STATE \parbox[t]{.9\linewidth}{
  Obtain the multiple homogeneous graphs $\hat G$ by the heterogeneous graph template.
} 
                %\hfill \(\triangleright\) Graph template
            \ENDFOR
            \FOR{each \(v \in V\), \(S \in G\)}
                \STATE \parbox[t]{.9\linewidth}{Obtain the prompt $P_{f}$ and the aggregation result 
$S_{v}$  of the subgraph by Eq. (\ref{6}).
}
            \ENDFOR
            \FOR{each \(i \in A + 1\)}
                \STATE \parbox[t]{.9\linewidth}{Obtain the prompt $P_{s}$ and the aggregation result 
$S_{v}$  of the subgraph by Eq. (\ref{eq.5}).
}
            \ENDFOR
            \FOR{each class \(c \in C\)}
                \STATE   \parbox[t]{.9\linewidth} {Obtain the mean of node/graph embedding vectors \(\tilde{\mathbf{s}}_{c} \) by Eq. (\ref{eq.3}).}
            \ENDFOR
            \FOR{each labeled pair \((x_i, y_i) \in \mathcal{M}\)} \label{algo:strat L}
                % \STATE \(Z_i \gets 0\)
                \FOR{each class \(c \in C\)}
                    \STATE  \parbox[t]{.9\linewidth} {Obtain the loss \( \mathcal{L}_{\text {down }}\) by Eq. (\ref{7}).}
                \ENDFOR
            \ENDFOR
            \STATE update \(P_f, P_s\) by minimizing \(\mathcal{L}_\text{down}.\)
        \ENDWHILE
        \RETURN \(P_f, P_s\)
    \end{algorithmic}
\end{algorithm}

\begin{algorithm}[!t]
    \caption{Prompt Evaluation And Pruning}
    \label{alg:algorithm 2}
    \renewcommand{\algorithmicrequire}{\textbf{Input:}}
    \renewcommand{\algorithmicensure}{\textbf{Output:}}
    \begin{algorithmic}[1]
        \REQUIRE \parbox[t]{.9\linewidth}{A heterogeneous graph \( {G} = \left (  {V},  {E} \right)  \) with multiple node types and edge types, labeled set \(\mathcal{M} = \left \{ \left ( x_{i}, y_{i} \right )\mid j = 1,2 \ldots \right \} \), class set \textit{C}, node type set \textit{A}, pre-trained GNN model \(f_{\Theta_{best}} \) \: takes a graph as input and outputs embeddings for the nodes \:\:\: in the graph, \(P_f\) and \(P_s\) after prompt tuning.} %%input
        \ENSURE \parbox[t]{.8\linewidth}{Pruned feature prompt \(\hat{P}_f\), pruned semantic prompt \(\hat{P}_s\).}  %%output
         
        \FOR{each \(\lambda_{i}\in \lambda\)} 
        
         \STATE \parbox[t]{.9\linewidth} {Obtain the loss \( \mathcal{L}_{\text {down }}\) by Eq. (\ref{7}).}
         
        \STATE \parbox[t]{.9\linewidth}{Obtain the importance score of semantic prompt \(  I_{p_{s}^{i}} \) by Eq.(\ref{ips}).}
        %\hfill \(\triangleright\)  Prompt Evaluation
            \IF {\(I_{p_{s}^{i}}<\delta\)}
                \STATE Prune the corresponding semantic prompt.       
                %\hfill \(\triangleright\)  Prompt Pruning
            \ELSE
                \STATE Retain the corresponding semantic prompt.
            \ENDIF
        \ENDFOR
        \FOR{each \(\eta_{j}\in \eta\)} 
         \STATE  \parbox[t]{.9\linewidth} {Obtain the loss \( \mathcal{L}_{\text {down }}\) by Eq. (\ref{7}).}
         
        \STATE \parbox[t]{.9\linewidth}{Obtain the importance score of semantic prompt \(  I_{p_{f}^{j}} \) by Eq.(\ref{ipf}).}
        
         \IF {\(I_{p_{f}^{j}}<\beta\)}  
                \STATE Prune the corresponding feature prompt blocks.
            \ELSE
                \STATE Retain the corresponding feature prompt blocks.
            \ENDIF
        \ENDFOR
        \STATE  \parbox[t]{.9\linewidth}{Obtain the semantic prompts \(\hat{P}_s\) and feature prompts \(\hat{P}_f\) after evaluation and pruning.}
        \RETURN \(\hat{P}_f,\hat{P}_s\)
    \end{algorithmic}
\end{algorithm}
\par\smallskip\noindent
{\bf Prompt Pruning.} 
\label{sec:4.3.2}
Overall, based on $I_{p_{s}^{i}}$  and $I_{p_{f}^{j}}$ we can tell which prompt tokens and prompt blocks are almost useless or have a negative effect on the model when processing the node classification task. Then we only need to set the value of \(\lambda_i\) or \(\eta_j\) to 0 to complete the prompt pruning operation. However, it should be noted that we need to determine how low the importance score will cause the prompt to be pruned. In order to better judge the importance scores, we first do z-score normalization on the importance scores. In addition, it is also important to set the threshold at which the importance score triggers Prompt Pruning. We set different thresholds for these two types of prompts. For semantic prompt $P_s$ we set a threshold \(\delta\) :
\begin{equation} \label{eq.12}
\lambda_i =\begin{cases}0&\text{if } I_{p_{s}^{i}}< \delta 
 \\
1&\text{if } I_{p_{s}^{i}}\ge  \delta 
\end{cases} \:
\end{equation}

We can then get a set of mask variables $\lambda$ corresponding to each token in the $P_s$, and with Eq. (\ref{eq p2p^}) we can get the pruned semantic prompt $\hat{P}_s$.
For $P_s$ we set a threshold of \(\beta\):
\begin{equation} \label{formula 14}
\eta _j =\begin{cases}0&\text{if } I_{p_{f}^{j}}< \beta
 \\
1&\text{if } I_{p_{f}^{j}}\ge  \beta
\end{cases} \:
\end{equation}

With Eq. (\ref{eq pf2pf^}) we can get the pruned feature prompt $\hat{P}_f$.
The choice of $\delta $ and $\beta$ will largely determine the degree of prompt pruning, and we decide the choice of thresholds according to the distribution of different graph prompts importance. Through Eqs (\ref{eq.12}) and (\ref{formula 14}) we can easily determine which prompts need to be pruned. For prompts that need to be pruned, we set the value of their corresponding mask variable to 0, so that the pruning operation can be completed without affecting other useful prompts. The processed prompts will be used as the initialization of the prompts during \emph{Retuning}.

\subsection{Retuning}
\label{sec:Retuning}
% According to the LTH theory, we also tune the pruned graph prompts to achieve the same prompting effect as the complete graph prompts, or even better. As shown in Fig. \ref{fig:model}(C), we adopt the same training strategy as prompt tuning and use heterogeneous graphs to complete NC task to tune prompts. The difference is that the prompts in this phase use pruned prompts to initialize the prompts, while the prompts in the prompt Tuning phase are randomly initialized. 
Inspired by the Lottery Ticket Hypothesis (LTH), we also fine-tune the pruned graph prompts to achieve the same or even better performance compared to using the complete graph prompts. As illustrated in Fig. \ref{fig:model}(C), we apply a training strategy similar to prompt tuning while utilizing heterogeneous graphs to complete the node classification task. The key distinction lies in the initialization of the prompts: in this stage, the prompts are initialized from the pruned prompts, whereas in the prompt \emph{Tuning} stage, they are initialized randomly. 
Specifically, we input a set of heterogeneous graphs \textit{G} into the GNN model that has been imported with frozen pre-trained parameters to obtain node feature embeddings. 
Then, use Eqs (\ref{6}) and (\ref{7}) to re-aggregate feature prompt \(\hat{P}_f\) and semantic prompt \(\hat{P}_s\) with the embedding of node features, formally as follows:
\begin{equation}  \label{eq:READOUT3}
 \operatorname{ReadOut}\left(\left\{\hat{P}_f  \odot {h}_{v} \mid v \in V(S)\right\}\right)\:
\end{equation}
\begin{equation}    \label{eq :READOUT4}
\operatorname{ReadOut}\left ( \left \{ \left ( 1\:+\:\hat{P}_{s}^{\text {i }} \right )  \odot\operatorname{ReadOut}\left( S^{i} \right)\mid S^{i}\in \hat{G}\left (  S\right )   \right \}  \right )\:
\end{equation}
the loss function is defined as : \begin{equation} \label{eq:Ldown2}
\small
 \mathcal{L}_{\text {down }}\left(\hat{P}_f,\hat{P}_s\right)=
-\sum_{\left(x_{i}, y_{i}\right) \in \mathcal{M}_{\text {down }}} \ln \frac{\exp \left(\frac{1}{\tau} \operatorname{sim}\left(\mathbf{s}_{x_{i}}, \tilde{\mathbf{s}}_{y_{i}}\right)\right)}{\sum_{c \in C} \exp \left(\frac{1}{\tau} \operatorname{sim}\left(\mathbf{s}_{x_{i}}, \tilde{\mathbf{s}}_{c}\right)\right)} \:
\end{equation}
where \(\mathcal{M}_{\text {down }}\) is a labeled training set, the subgraph embedding \(\mathbf{s}_{x_{i}} \) is generated based on feature prompt \(\hat{P}_f\) and semantic prompt \(\hat{P}_s\) by using Eqs. (\ref{eq:READOUT3}) and (\ref{eq :READOUT4}). 
Through this loss function, we can learn more effective prompts with parameters. 
In addition, pruned prompts are shorter, have fewer parameters than the original prompts, and are already well-initialized.
The \emph{Retuning} phase trains for fewer epochs than the prompt tuning loss function to reach convergence, allowing the retained positive prompts to receive more thorough tuning. 

% Therefore, in the Retuning phase, our purpose is \textit{Make Fewer Graph Prompts More powerful.} 

\subsection{Algorithm Design and Analysis}
\label{sec:algorithm}
In this section, we will elaborate on the time and space complexities of the \alg~framework.
\par\smallskip\noindent
\textbf{Prompt Tuning} (Alg. \ref{alg:algorithm 1}). 
The initialization and embedding computation have a time complexity of \(O(|V| + |E|)\). The outer while loop runs \(A + 1\) times, with each iteration involving traversing nodes and edges, resulting in a time complexity of \(O(|V| \cdot |S|)\). 
The mean calculation for each class and the computation for each labeled pair have time complexities of \(O(|C|)\) and \(O(|C| \cdot |\mathcal{M}|)\), respectively. 
Therefore, the total time complexity is \(O((A + 1) \cdot |V| \cdot |S| + |C| \cdot |\mathcal{M}|)\). 
The space complexity for initialization and embedding storage is \(O(|V|)\). During the while loop, storing intermediate results and traversing nodes and edges have a space complexity of \(O(|V| \cdot |S|)\). 
Storing the mean vector for each class and the labeled pairs has space complexities of \(O(|C|)\) and \(O(|\mathcal{M}|)\), respectively. 
Consequently, the total space complexity is \(O(|V| + |V| \cdot |S| + |C| + |\mathcal{M}|)\).

% Algorithm \ref{alg:algorithm 1} illustrates the Prompt Tuning framework of \alg. In line 1, we initialize two types of prompts, and the loss function value is set to 0. In line 2 we import the GNN parameters obtained from pre-training into the present GNN model, and input the heterogeneous graph data to get the corresponding embedding. Then we start prompt tuning using the NC task. Lines 4-5 use graph templates for a heterogeneous graph to get a set of homogeneous graphs. Lines 6-9 are the embeddings of the nodes with the two prompt aggregation operations. lines 10-11 compute the embeddings of the node category prototypes. Lines 12-17 show the steps to optimize the model using the loss function. The trained two types of prompts are finally obtained.

\par\smallskip\noindent
\textbf{Prompt Evaluation and Pruning} (Alg. \ref{alg:algorithm 2}).
The initialization step has a time complexity of \(O(1)\). 
The first for-loop iterates over \(|\lambda|\) elements, and each iteration involves computing \(\mathcal{L}_{\text{down}}\) by invoking lines 12-16 of Alg. \ref{alg:algorithm 1}, which has a time complexity of \(O((A + 1) \cdot |V| \cdot |S| + |C| \cdot |\mathcal{M}|)\), plus an expectation calculation with time complexity of \(O(|\mathcal{M}|)\). 
The total time complexity of this loop is \(O(|\lambda| \cdot ((A + 1) \cdot |V| \cdot |S| + |C| \cdot |\mathcal{M}|))\). 
Similarly, the second for-loop iterates over \(|\eta|\) elements, leading to the same time complexity of \(O(|\eta| \cdot ((A + 1) \cdot |V| \cdot |S| + |C| \cdot |\mathcal{M}|))\). 
Updating and returning the prompts have a time complexity of \(O(1)\). 
Therefore, the total time complexity of Alg. \ref{alg:algorithm 2} is \(O((|\lambda| + |\eta|) \cdot ((A + 1) \cdot |V| \cdot |S| + |C| \cdot |\mathcal{M}|))\). 
For space complexity, initializing \(P_f\) and \(P_s\) requires \(O(1)\) space. The for-loops require space for storing \(|\lambda|\) and \(|\eta|\) elements, respectively. 
Storing intermediate results from Alg. \ref{alg:algorithm 1} incurs a space complexity of \(O(|V| + |V| \cdot |S| + |C| + |\mathcal{M}|)\). 
Thus, the total space complexity is \(O(|\lambda| + |\eta| + |V| + |V| \cdot |S| + |C| + |\mathcal{M}|)\).

% Algorithm \ref{alg:algorithm 2} illustrates the prompt evaluation and pruning framework of \alg. In addition to what is required by Algorithm \ref{alg:algorithm 1}, we also need the two types of prompts trained by Algorithm \ref{alg:algorithm 1} to be used as the initialization values for the two types of prompts in line 1. Starting from the line 2, we will evaluate these two prompts separately. The first is the semantic prompt, and each part of it will be evaluated individually. In line 3, we will use lines \ref{algo:strat L}-\ref{algo:end L} of Algorithm \ref{alg:algorithm 1} to calculate the loss function. In line 4, we will use Eq. (\ref{ips}) to calculate the importance scores of semantic prompt. Lines 5-8 will involve judging the importance scores, pruning the prompt token if the score is less than the threshold, and keeping it if it exceeds the threshold. Lines 9-11 will involve calculating the importance scores for each block of the feature prompt after dividing. Eq. (\ref{ipf}) will be used to calculate the importance scores, and lines 12-15 will involve judging whether to prune or retune  based on the scores. Finally, we will obtain the pruned two types of prompts.

\par\smallskip\noindent
\textbf{Retuning.} 
The retuning is essentially the same as that of Alg. \ref{alg:algorithm 1}, with the difference being the use of different prompt values during adjustment. 
When retuning, the final prompt values obtained from Alg. \ref{alg:algorithm 2} are used for initialization.

\par\smallskip\noindent

%% file: B-5-Experiments-A.tex
% =============================================
% # Ⅴ. Experiments #
% =============================================
\section{Experiments}
\label{sec:exp}
In this section, we validate and evaluate the performance of our model through extensive experiments conducted on three real-world datasets.
We are committed to solving the following three problems ({\bf RQs}):
\begin{itemize}[leftmargin=*]
    \item {\bf RQ1}: How does our model outperform the state-of-the-art in the node classification task?
    \item {\bf RQ2}: How do the importance levels of various graph prompts vary?
    % Do all graph prompts play the same role in downstream tasks? 
    \item {\bf RQ3}: Can overall performance be improved by pruning prompts with low importance scores?
    \item {\bf RQ4}: Is \alg~sensitive to hyperparameters?
\end{itemize}

\begin{table}[!t]
    \begin{center}
    \caption{Characteristics of the studied datasets.}
    \setlength{\tabcolsep}{4pt}
    \label{tab:datasets}
    \scalebox{1.1}{
    \begin{tabular}{ c | c c c c c c }
        \toprule[0.9pt]
      \multirow{2}{*}{\textbf{Dataset}} &\multirow{2}{*}{\textbf{Nodes}}& \textbf{Node}  &\multirow{2}{*}{\textbf{Edges}} &\textbf{Edge} & \textbf{Target} &\multirow{2}{*}{\textbf{Classes}}\\
        &  & \textbf{Types} &  &\textbf{Types}&\textbf{Types}\\
        \midrule
        ACM      &10,942  &4 &547,872   &8  &paper  &3\\
        DBLP     &26,128  &4 &239,566   &6  &author &4\\ 
        Freebase &180,098 &8 &1,057,688 &36 &book   &7\\
        \bottomrule[0.9pt]
    \end{tabular}
    }
    \end{center}
\end{table}

\subsection{Experiment Settings} 
\label{sec:datasets}
\par\smallskip\noindent
{\bf Datasets.}
% We complete experiments on three benchmark datasets. We apply the same settings from prior studies\cite{lv2021we} to three datasets.
We apply the same settings from prior studies to complete experiments on three benchmark datasets.
(1) {\bf ACM}~\cite{wang2019heterogeneous} is utilized as a citation network, encompassing papers from five conferences further categorized into three distinct domains: Database, Wireless Communication, and Data Mining. 
(2) {\bf DBLP}~\cite{lv2021we} is an extensive bibliographic database hosting research papers and computer science proceedings. 
(3) {\bf Freebase}~\cite{bollacker2008freebase} is a comprehensive structured repository designed to capture and organize diverse information about entities and their global relationships. 
% We used the same original data set from the HGB library (Simple-HGN)~\cite{lv2021we} and a similar data sampling process as HGPrompt. 
Table~\ref{tab:datasets} summarizes the datasets we used.

\begin{table*}
    \centering
    \caption{Node classification results (\%) and their standard deviations on three real-world datasets under the one-shot setting. Missing values ("-") indicate that the model is running out of memory on large graphs.}
  \setlength{\tabcolsep}{9pt}
  \label{tab:SOTA}
  \scalebox{1.2}{
  \begin{tabular}{l|c cc cc cc}
    \toprule[0.9pt]
    \midrule
       % &\multicolumn{6}{c}{Node Classification} \\
       % \midrule \multicolumn{1}{l|} {Methods}\multirow{2}{*}{Methods}
       \multirow{2}{*}{\textbf{Methods}} &\multicolumn{2}{c}{\textbf{ACM}}  &\multicolumn{2}{c}{\textbf{DBLP}} &\multicolumn{2}{c}{\textbf{Freebase}} \\
        &Micro-F(\%)& Macro-F(\%) & Micro-F(\%)& Macro-F(\%) & Micro-F(\%)& Macro-F(\%)\\
    \midrule
  \multicolumn{1}{l|}{GCN} &  50.98±0.13  & 45.02±0.17&48.45±0.16& 44.74±0.19
  &19.91±0.12&15.73±0.09\\
  \multicolumn{1}{l|}{GAT} & 39.85±0.10& 29.33±0.15 & 65.16±0.12& 63.03±0.16
  & 22.1±0.11 &\pmb{20.28±0.08}\\
  
  \midrule
  \multicolumn{1}{l|}{Simple-HGN} &  44.11±0.11  & 36.31±0.16 &62.28±0.14& 59.81±0.15
  &21.28±0.09 &\underline{18.94±0.08}\\
  \multicolumn{1}{l|}{HAN} & 62.14±0.08& 57.53±0.11 & 62.42±0.19& 60.76±0.20
  & 16.38±0.11& 9.80±0.07\\

  \midrule
  \multicolumn{1}{l|}{DGI} &  62.88±0.06  & 50.38±0.09 &62.74±0.11& 62.37±0.15
  & -&- \\
  \multicolumn{1}{l|}{GRAPHCL} & 59.98±0.38& 22.07±0.13& 65.03±0.07 & 53.64±0.10
  & -& -\\
    \midrule
  \multicolumn{1}{l|}{CPT-HG} &62.57±0.15   &  58.47±0.20 & 70.63±0.13&  67.03±0.13
  &  19.00±0.09 & 17.46±0.06  \\
  \multicolumn{1}{l|}{HECO} & 63.32±0.15  &  59.13±0.17 & 70.74±0.12  & 67.63±0.12
  &19.42±0.08 &18.17±0.06 \\
    \midrule
  \multicolumn{1}{l|}{GPPT} &  54.89±0.11  & 51.67±0.12 &61.47±0.09& 63.23±0.11
  & -&- \\
  \multicolumn{1}{l|}{GraphPrompt} & 62.67±0.18& 58.86±0.17 & 65.16±0.12& 63.03±0.16
  &- & -\\
    \multicolumn{1}{l|}{All in One} & 56.05±0.05& 47.53±0.08 & 54.12±0.16& 27.70±0.07
  & - &- \\
  \multicolumn{1}{l|}{HGPrompt} & \underline{71.60±0.12}& \underline{68.14±0.14} & \underline{79.25±0.09}&\underline{78.00±0.09} 
  &\underline{23.67±0.05} & 14.60±0.02\\
      \midrule
  \multicolumn{1}{l|}{\alg~(Ours)} &  \pmb{72.98±0.14}  & \pmb{71.04±0.16} &\pmb{82.57±0.08}& \pmb{81.31±0.09}
  &\pmb{25.32±0.06}  & 16.13±0.02   \\
     \midrule
    \bottomrule[0.9pt]
  \end{tabular}
  }
\end{table*}

\par\smallskip\noindent
{\bf Baselines.} 
\label{sec:baselines}
We evaluated our model based on state-of-the-art technology in the following five main categories as shown below: 
\begin{itemize}[leftmargin=*]
    \item {\bf End-to-end homogeneous graph neural networks (HmGNNs)}: It is a graph neural network that learns node representations on homogeneous graphs to capture their structural relationships, \emph{i.e,} GCN~\cite{kipf2016semi} and GAT~\cite{velickovic2017graph}.
    \item {\bf End-to-end heterogeneous graph neural networks (HetGNNs)}: It is a graph neural network that processes heterogeneous graphs by learning diverse node and edge representations, \emph{i.e,} Simple-HGN~\cite{lv2021we} and HAN~\cite{wang2019heterogeneous}
    \item {\bf Homogeneous Graph pre-training models (HmGPTs)}: It is a pre-training approach that leverages homogeneous graph data to enhance the learning of node and graph-level representations, \emph{i.e,} DGI~\cite{velivckovic2018deep} and GraphCL~\cite{you2021graph}.
    \item {\bf Heterogeneous Graph pre-training models (HetGPTs)}: It is a pre-training approach that utilizes heterogeneous graph data to improve the learning of diverse node and edge representations, \emph{i.e,} CPT-HG~\cite{jiang2021contrastive} and HeCo~\cite{wang2021self}.
    \item {\bf Graph Prompt Models (GPs)}: It transforms downstream tasks into pre-training tasks or narrows the gap between downstream tasks and pre-training tasks, \emph{i.e,} GPPT~\cite{sun2022gppt}, GraphPrompt~\cite{liu2023graphprompt}, All in One~\cite{sun2023all} and HGPrompt~\cite{yu2024hgprompt}.
\end{itemize}

% (1) \textit{End-to-end homogeneous graph neural networks}: GCN~\cite{kipf2016semi} and GAT~\cite{velickovic2017graph}. These homogeneous GNNs utilize neighborhood aggregation as their core operation, gathering messages from neighboring nodes iteratively in an end-to-end fashion.  
% (2)\textit{ End-to-end heterogeneous graph neural networks(HGNNs)}: Simple-HGN and HAN~\cite{wang2019heterogeneous}. HGNNs differ from conventional GNNs by incorporating heterogeneity through heterogeneous neighborhood aggregation, considering edge types or meta-paths. 
% (3) \textit{Homogeneous Graph pre-training models}: DGI~\cite{velivckovic2018deep} and GraphCL~\cite{you2021graph} follow the "pre-training, fine-tuning" paradigm, first completing pre-training on the GNN model through self-supervised learning of the graph structure, and then fine-tuning the GNN model weights on downstream tasks.
% (4) \textit{Heterogeneous Graph pre-training models}: CPT-HG~\cite{jiang2021contrastive} and HeCo~\cite{wang2021self} also follow the "pre-training, fine-tuning" paradigm, but their difference from Homogeneous Graph pre-training models is that in the pre-training stage, heterogeneous graph is used to complete heterogeneous tasks to learn heterogeneous information.
% (5) \textit{Graph Prompt Models}: GPPT~\cite{sun2022gppt}, GraphPrompt and HGPrompt. They all reformulate downstream tasks into pre-training tasks or close the gap between downstream tasks and pre-training tasks.

\indent Consistent with the pre-training stage settings of GPPT, GraphPrompt and HGPrompt, we use label-free graphs to complete the link prediction task.

\par\smallskip\noindent
{\bf Hyperparameter settings.} 
\label{sec: parameter settings}
We choose node classification as the downstream task to verify the effectiveness of our framework, which follows a \textit{k}-shot learning~\cite{wang2020graph} setting. The task construction process will be further elaborated when the results of the task are presented in Section \ref{section 5.2}. We adopt Micro-F and Macro-F as evaluation metrics.

\indent In the settings of our model. we employ a 2-layer GCN as the base model and set the hidden dimensions as 64. We set the optimal number of blocks\textit{ t} for feature prompts to 16. For $P_s$ we set the threshold of \(\beta =0.4 \) and for semantic prompt $P_s$ we set the threshold \(\delta = 0.6\). 
For the baseline models in HmGNNs and HetGNNs, we use the Heterogeneous Graph Benchmark (HGB)~\cite{lv2021we} library. Taking inspiration from ProG~\cite{sun2023graph}, we adopted similar parameter settings and dataset preprocessing approaches for other models. The downstream tasks of all baseline models are node classification in \textit{k}-shot settings. For all models, we are based on the author's code and default settings and the models in the above libraries are based on the code and default settings in the library.

\begin{figure}[h] 
    \centering
    \hspace*{-0.3cm}\includegraphics[width=0.50\textwidth]{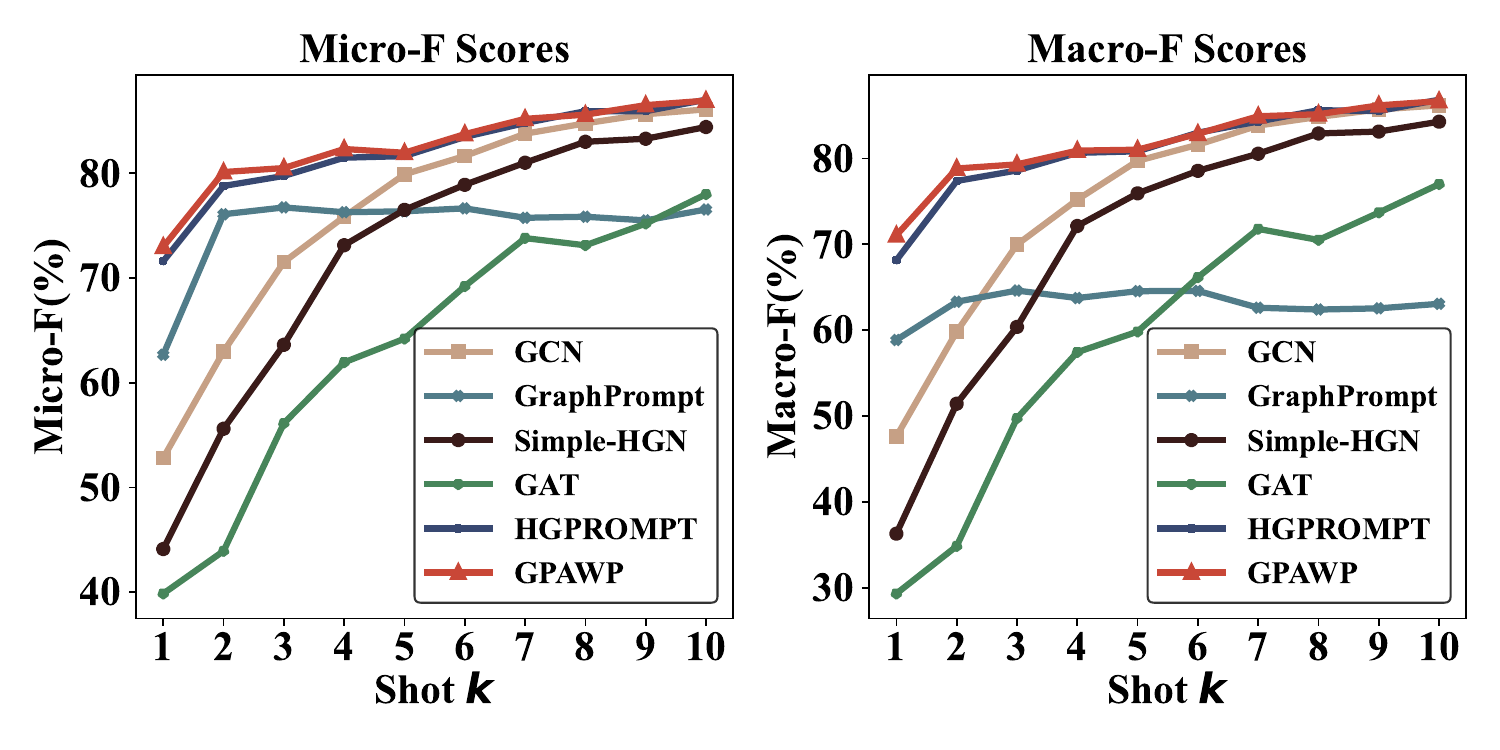}
    \caption{ The impact of shot nums on the NC task of the ACM dataset.}
    \label{fig:shot-graph}
\end{figure}

%We initialize embeddings at size 64 with Gaussian parameters (mean 0, standard deviation 0.1). MFE settings include learning rate of $5e^{-4}$, $L_{2}$ regularization weight of $1e^{-4}$, and dropout ratio of 0.2, with loss weight $\beta$ of 1. 

\input{B-5-Experiments-RQ1}

\input{B-5-Experiments-RQ2}

\input{B-5-Experiments-RQ3}

\input{B-5-Experiments-RQ4}

%% file: B-5-Experiments-RQ1.tex
\subsection{Overall Performance (RQ1)}
\label{5.2}
% Performance Comparison

% \begin{figure}[!t]
%     \centering
%      \begin{subfigure}[!t]{0.47\linewidth}
%         \includegraphics[width=\textwidth]{Exp-Figs/macro.pdf} 
%         \caption{\label{fig:CH1}
%         Challenge \uppercase\expandafter{\romannumeral1}
%         }
%     \end{subfigure} 
%     \begin{subfigure}[!t]{0.43\linewidth}
%         \includegraphics[width=\textwidth]{Exp-Figs/micro.pdf}
%         \caption{\label{fig:CH2}
%         Challenge \uppercase\expandafter{\romannumeral2}
%         }
%     \end{subfigure}
%     \caption{
%     \label{fig:violin_CAR}
%     Illustration of Challenges. 
%     \emph{Left}: Trend curves of confidence changes for different feature information under the german dataset.
%     \emph{Right}: Trends of cumulative accuracy rate (CAR) under the hapt dataset. $\rho$ represents the noise rate of the label from the original class to another class.
%     }
% \end{figure}

% \begin{figure}[!t]
%     \centering
%     \subfigure[Coat]{
%     \includegraphics[width=0.47\linewidth]{Exp-Figs/macro.pdf}
%     }
%     \subfigure[Movielens-1m]{
%     \includegraphics[width=0.47\linewidth]{Exp-Figs/micro.pdf}}
%     \caption{Trend in layer-to-ego similarity changes
%     % Trend of changes in similarity between various layers and the ego layer
%     % Trend of changes in \alg~ each layer 
%     (Ego layer is adaptive weights and 1-3 are similarity weights). }
%     \label{fig:trend}
% \end{figure}

\label{sec:performance comparison} \label{section 5.2}
In this section, we provide a comprehensive comparison results of \alg~and five types of methods on three datasets in the 1-shot setting (Table \ref{tab:SOTA}). 
We then evaluate the six models on the ACM dataset with different shot settings (Fig. \ref{fig:shot-graph}). Following the conventional \textit{k}-shot framework~\cite{liu2021relative}, we randomly generate 100 one-shot tasks for training and validation of the model. For the NC task (\emph{i.e.}, in each task, we randomly sample one node per class) for training and validation, respectively.

% From Table \ref{tab:SOTA}, it is clear that \alg~ achieves the best results on all three datasets, with 1.38\% improvement in Micro-F and 2.9\% improvement in Macro-F on the ACM dataset compared to the next best HGPrompt, and 3.32\% improvement in Micro-F and 3.31\% improvement in Macro-F on the DBLP dataset compared to HGPrompt. Only on the Freebase dataset the Macro-F values did not yield good results, but there was still a 1.65\% improvement in Macro-F compared to HGPrompt (the model for which we performed graph prompt pruning). In addition, we also analyze the reason why GPPT, All in One and GraphPrompt do not perform well, which may be that they are specifically designed for homogeneous graphs. The datasets used in this paper are all heterogeneous graphs, and the heterogeneous graph data do not match the homogeneous graph prompts of these models, so they do not play the role of graph prompts well.

Table \ref{tab:SOTA} shows that \alg~achieves the best results on all three datasets. It shows a 1.38\% improvement in Micro-F and a 2.9\% improvement in Macro-F on the ACM dataset compared to the next best performer, HGPrompt. Similarly, on the DBLP dataset, \alg~ shows a 3.32\% improvement in Micro-F and a 3.31\% improvement in Macro-F compared to HGPrompt. The only exception is the Freebase dataset, where the Macro-F values did not yield satisfactory results. We analyzed that if the model is particularly optimized for a minority of target node types, resulting in very high F1 scores for these types, but does not apply similar optimizations to the majority of other node types, this imbalance may cause the macro-F score to be lower while the micro-F score remains high (\emph{e.g}. in the Freebase dataset, the BOOK type is the target node of the classification task is much less common than other node types).
Specifically, we analyze why GPPT, All in One, and GraphPrompt underperform, possibly due to their specific design for homogeneous graphs. Since the datasets used in this study are all heterogeneous, the prompts designed for homogeneous graphs do not fit well with the heterogeneous graph data and thus do not serve their intended purpose.

To answer \textbf{RQ1}, we conducted experiments on the ACM dataset with different numbers of shots and compared five benchmark models (Fig. \ref{fig:shot-graph}). The results show that \alg~achieved the best performance when the number of shots was between 1 and 5. As the number of shots increases, both Micro-F and Macro-F scores tend to stabilize. This is because, with the increase in shots, the training samples also increase, requiring the prompts to carry more information. However, a small number of prompts cannot carry a large amount of information. Our model outperforms the state-of-the-art in the node classification task by leveraging the effectiveness of positive prompts with limited data, maintaining high performance even with fewer training samples. This efficiency and adaptability make \alg~superior in scenarios where traditional GNN models may struggle due to insufficient data.

% To answer \textbf{RQ1}, we set up different number of shots on the ACM dataset and conduct experiments on five benchmark models for comparison (Fig. \ref{fig:shot-graph}). It can be seen that our models achieve optimal results when the number of shots is in the range of 1-5, which suggests that a small number of prompts can have a great effect when a small amount of training data is available, and too many prompts are likely to lead to negative effects. With this we can answer \textbf{RQ1}, compared to traditional GNN models graph prompt can effectively use the a priori knowledge learned from the pre-training task to accomplish the NC task in the downstream task, and compared to the normal graph prompt model our method uses less graph prompts to provide more effective prompting information, and therefore can outperform the baseline model. When the number of shots is 5-8, our model has similar results with HGPrompt, and we analyze the results because a small number of prompts can not provide enough prompt information to complete the NC task with the increase of training data.

%% file: B-5-Experiments-RQ2.tex
\begin{figure}[!t]
    \centering
    % \hspace*{-0.2cm}
    \includegraphics[width=0.50\textwidth]{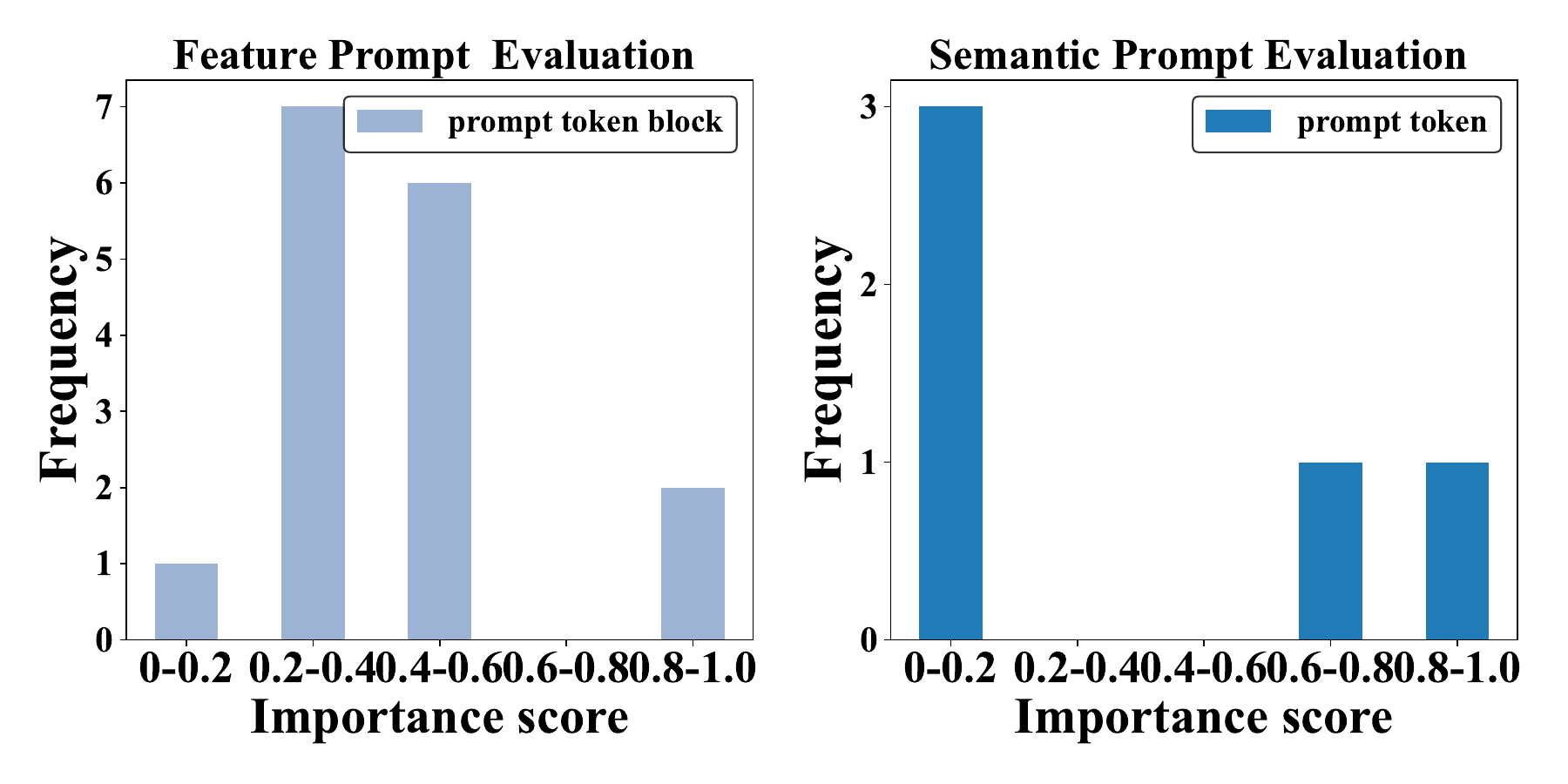}
    \caption{The importance scores for feature prompt token blocks and semantic prompt tokens on the ACM dataset.}
    \label{fig:trend}
\end{figure}

\subsection{Efficiency Comparison (RQ2)} % Ablation and effectiveness studies
\label{ablation and effectiveness studies}
% \subsubsection{Parameter Efficiency.}

\par\smallskip\noindent
\textbf{Parameter Efficiency.}
We empirically demonstrate that \alg~is superior to other models in parameter efficiency. Specifically, we analyze the parameter requirements for optimization in downstream tasks for different models (Table \ref{tab3}). For the NC task on three datasets, we examine the number of parameters to optimize for each model. First, GCN and Simple-HGN, which use an end-to-end approach, update nearly all parameters in each training iteration, resulting in a large parameter volume. In contrast, models such as DGI and CPT-HG follow a "pre-train, fine-tune" paradigm, updating only the parameters of the downstream task classifiers, reducing the number of parameters and improving training efficiency. Finally, graph prompt methods such as GraphPrompt and HGPrompt only update prompt parameters, resulting in fewer parameters. Notably, our model prunes graph prompts, reducing the number of parameters to be updated in each training round, making it more parameter efficient.

\par\smallskip\noindent
\textbf{Efficiency of Tuning Time.} 
We demonstrate through experiments that \alg~outperforms other models in terms of time efficiency. Specifically, we perform node classification tasks during the graph prompt tuning phase on the DBLP dataset and compare the training and testing times per epoch for several multi-class models (Table \ref{tab4}). From Table \ref{tab4}, we can see that the graph prompt models are faster to train (measured in seconds per epoch) compared to the regular graph neural network models. This is because the graph prompt models only need to tune the parameters in the graph prompt part of the model. Our \alg~has the fastest training speed of these models because it needs to tune fewer graph prompt parameters.

% We further empirically demonstrate the superior training efficiency of \alg~ compared to other models, particularly GNN-based models. Specifically, We delve deeper into analyzing the parameter requirements for optimization in downstream tasks across a range of representative models (Table \ref{tab3}). We analyze the number of parameters to be optimized for each model in the NC task on the three datasets. 
% Firstly, GCN and Simple-HGN, two models employing an end-to-end approach, update nearly all parameters at each training iteration, resulting in a considerable parameter volume. Then there are models like DGI and CPT-HG that use the ''pre-train, fine-tune'' paradigm, where they only have the parameters of the downstream task classifiers for each update iteration, so the number of parameters is drastically reduced and training is more efficient compared to end-to-end approaches. Finally, there are the graph prompt methods GraphPrompt and HGPrompt, which only require updates to the parameters of the prompt and therefore have fewer parameters. It is worth noting that our model can prune the cues, and thus has less number of parameters to update at each round of training compared to them, making it more parameter effective.

To answer \textbf{RQ2}, we conducted an additional experiment (Fig. \ref{fig:trend}). The aim was to calculate importance scores for each block in the feature prompt and each token in the semantic prompt. 
As depicted in Fig. \ref{fig:trend}, the distribution of scores for blocks in the feature prompt and tokens in the semantic prompt is sparse, and these distributions differ between the two types of prompts. 
This suggests that there are indeed significant differences in the importance of different graph prompts. 
Lower scores indicate that the corresponding prompt token or block has a reduced or potentially negative impact on downstream tasks. 
Based on the different distributions of importance scores, we can prune the negative prompts in both types of prompts (those with importance scores lower than their respective thresholds), thereby enhancing the parameter efficiency of the model. 
The selection of thresholds (\(\beta = 0.4\) for feature prompts, \(\delta =0.6\) for semantic prompts) is driven by the distribution patterns of importance scores and task-specific objectives. 
As shown in Fig. \ref{fig:trend}, feature prompt blocks exhibit a left-skewed distribution (densely clustered in 0–0.4), where pruning low-score blocks (50\%) removes redundancies while preserving moderate-impact features (0.4–1.0). 
For semantic prompts with a right-skewed distribution (scores >0.6 dominate), (\(\beta = 0.4\) prunes only sparse low-score tokens (<10\%) to protect structural semantics. Increasing (\(\beta\) to 0.5 marginally improves accuracy at higher computational cost, while lowering it to 0.3 causes significant performance degradation; similarly, \(\delta >0.7\) risks over-pruning critical mid-high score tokens, and \(\delta<0.5\) introduces noise. These thresholds, validated by distribution alignment and robustness tests, optimally balance pruning efficiency and task performance.

% To answer \textbf{RQ2}, we also conducted the experiment  (Fig. \ref{fig:trend}). This experiment is to calculate the importance score of each block in the feature prompt and the importance score of each token in the semantic prompt. As shown in Fig. \ref{fig:trend}, we can clearly see that the distribution of scores for each block in the feature prompt is sparse, i.e., there is indeed a negative prompt block, which plays a less prompting role in the downstream task than the positive prompt. The same is true for the semantic prompt, which indicates that the subgraphs corresponding to different prompt tokens play different roles in the downstream tasks, and the lower the scores are, the smaller or even negative roles they play in the downstream tasks.

% select 

% \begin{figure}[!t]
%     \centering
%     \subfigure[Feature prompt blocks]{
%     \includegraphics[width=0.48\linewidth]{Exp-Figs/DBLP_prompt_f_imp_score.pdf}
%     }
%     \subfigure[Semantic prompts]{
%     \includegraphics[width=0.48\linewidth]{Exp-Figs/DBLP_prompt_s_imp_score.pdf}}
%     \caption{The importance scores for semantic prompts and feature prompt blocks}
%     \label{fig:trend}
% \end{figure}

\begin{figure}[!t]
    \centering
    % \hspace*{-0.2cm}
    \includegraphics[width=0.5\textwidth]{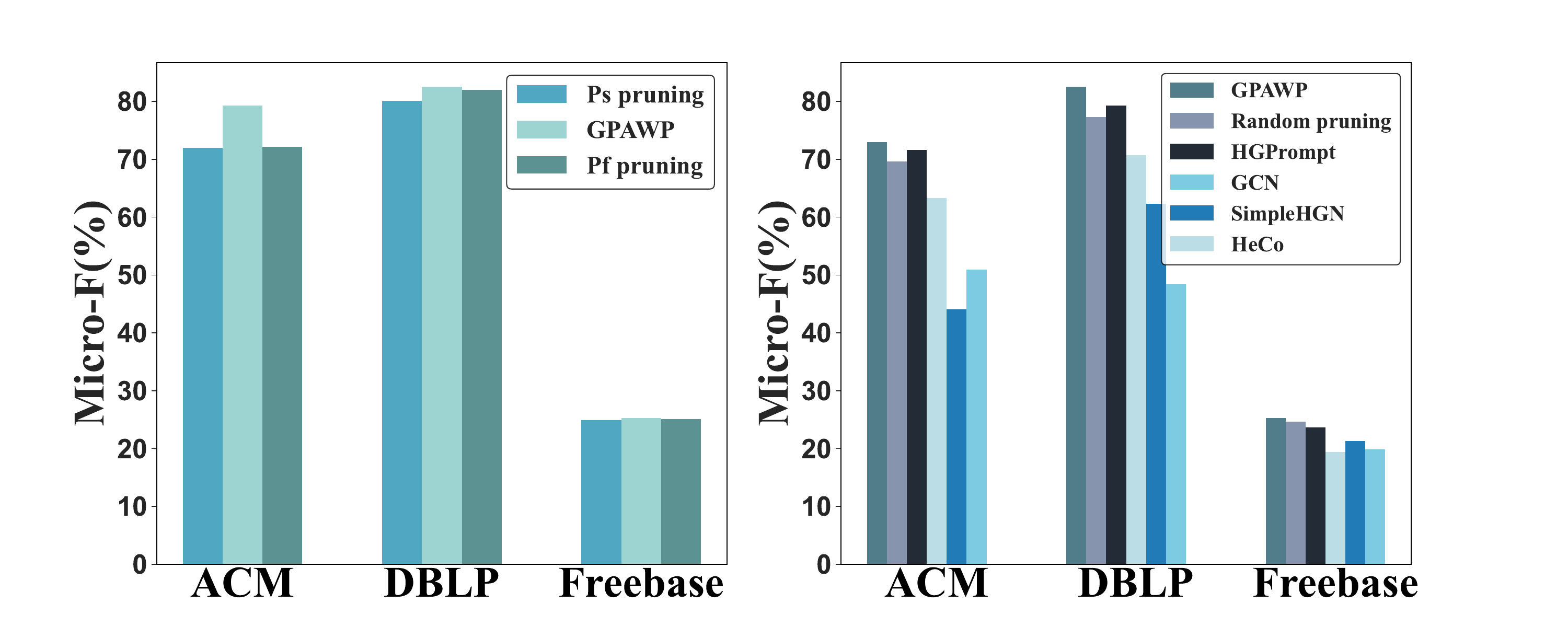}
    \caption{Comparative analysis of model variants with component variations in disruption experimentation and comparative evaluation of model variants and baseline models.}
    \label{fig:difAgg}
\end{figure}

\begin{table*}[!t]
\begin{center}
    \caption{Variants used in ablation study, and corresponding results in Micro-F (\%) and Macro-F(\%).}
    \label{tab5}
    \setlength{\tabcolsep}{6pt}
    \scalebox{1}{
    \begin{tabular}{ c | c  c  c | c  c c c c c}
    \toprule%[0.9pt]
    \midrule
        \multirow{2}{*}{\textbf{Methods}} & \multirow{2}{*}{\textbf{Tuning}} & \multirow{2}{*}{\textbf{Evaluation \& Pruning}}&\multirow{2}{*}{\textbf{Retuning}}  & \multicolumn{2}{c}{\textbf{ACM}} & \multicolumn{2}{c}{\textbf{DBLP}} & \multicolumn{2}{c}{\textbf{Freebase}}\\
          & &  &  & Micro-F(\%) & Macro-F(\%)  & Micro-F(\%) & Macro-F(\%) & Micro-F(\%) & Macro-F(\%)\\        
    \midrule
    \emph{w/o} REP &\(\surd \) & \(\times\) & $\times $&72.10&69.25 &79.55 &78.45&23.67&14.6 \\
    \emph{w/o} R &\(\surd \) & \(\surd\) & $\times $&71.07&68.03&79.33&78.17&23.25&14.95 \\
    \emph{w/o} EP &\(\surd \) & $\times$ & \(\surd\)&73.48&70.90 &81.32&80.42&24.66&15.26 \\
    \alg~&\(\surd \) & \(\surd\) & \(\surd\)& \pmb{74.07}& \pmb{71.52}&\pmb{82.57}&\pmb{81.31}&\pmb{25.32}&\pmb{16.13} \\
    \midrule
    \bottomrule%[0.9pt]
    \end{tabular}    
    }
\end{center}
\end{table*}

%% file: B-5-Experiments-RQ3.tex
\subsection{Ablation study (RQ3)}
\label{sec:speaker embedding analysis}
% A model analysis was conducted to examine the impact of each component of the model. We modified our model based on variants of component designs related to \textit{Prompt Evaluation and Pruning} (Fig. \ref{fig:difAgg}). We completed the NC task on each of the three datasets under the one-shot setting for comparison. "ps pruning" denotes a model that evaluates and prunes only semantic prompts, "pf pruning" indicates a model that evaluates and prunes only feature prompts, and the number of blocks is set to 16. \alg~ represents the complete model. It can be seen that the evaluation and pruning of feature prompt is higher for the overall model enhancement than the evaluation and pruning of semantic prompt, but both have some enhancement for the model. 

Ablation study was conducted to examine the impact of each component of the model. 
To investigate the impact of each component in the model, we conducted ablation experiments on the three key components of the model (Table \ref{tab5}). 
It is important to note that since the tuning component is a crucial part of the graph prompt model, all variants retained this component.
\textbf{Variant 1:} without \emph{Retuning} and \emph{Evaluation \& Pruning} component (\emph{w/o} REP). 
\textbf{Variant 2:} without \emph{Retuning} component (\emph{w/o} R), and \textbf{Variant 3:} without \emph{Evaluation \& Pruning} component (\emph{w/o} EP). 
The results in Table \ref{tab5} indicate the following: (1) The absence of \emph{Retuning} on the graph prompts after \emph{Evaluation \& Pruning} significantly reduces model performance. 
This is because the prompt information carried by the graph prompts decreases with the pruning operation, making it necessary to increase the remaining prompt information through \emph{Retuning}. 
(2) The absence of the \emph{Evaluation \& Pruning} component also somewhat reduces model performance, highlighting the necessity of evaluating and pruning the graph prompts.

In addition, We also modified our model based on variants of the component designs related to prompt evaluation and pruning (as shown in the left subplot of Fig. \ref{fig:difAgg}). 
We completed the NC task on each of the three datasets under the one-shot setting for comparison. 
"Ps pruning" represents a model that evaluates and prunes only semantic prompt tokens, and "Pf pruning" represents a model that evaluates and prunes only feature prompt blocks (the number of blocks \textit{t} is set to 16). 
\alg~represents the complete model. It is clear that evaluating and pruning the feature prompt improves the overall model more than evaluating and pruning the semantic prompt. However, both approaches have a positive impact on the model.

To answer\textbf{ RQ3}, we also conducted an experiment (Figure \ref{fig:difAgg}) using our model \alg~and a variant of \alg~called "Random pruning" (randomly selecting the same number of prompt tokens and prompt token blocks as \alg~ for pruning without evaluation) and compared it with four different types of baseline models. 
It can be seen that the effect of random pruning on model improvement is unstable, \emph{e.g.}, in the ACM and DBLP datasets the "Random pruning" variant of the model is even worse than the baseline model HGPrompt. While in the ACM, DBLP, and Freebase datasets, the effect of our model \alg~ is much higher than that of all baseline models. Thus, the model effect can be improved by pruning the prompts with lower importance scores.
% To answer \textbf{RQ3}, we also performed the experiment(Fig. \ref{fig:difAgg}), using our model \alg~ and a variant of \alg~ called "Random Pruning" (randomly selecting the same number of prompt tokens and prompt token blocks as \alg~ for pruning) and comparing it with four different types of baseline models. It can be found that the effect of random pruning on model enhancement is unstable, such as in the ACM and DBLP datasets "Random pruning" variant of the model is even worse than the baseline model HGPrompt, while in the ACM, DBLP and Freebase datasets, the effect of our model \alg~ is much higher than that of all the baseline models. Thus the model effect can be improved by pruning the prompts with lower importance scores.

\begin{table}[!t]
    \begin{center}
    \caption{Comparison of parameters for downstream node classification.}
    \label{tab3}
    \setlength{\tabcolsep}{7pt}
    \scalebox{1.2}{
        \begin{tabular}{c | c c c }
        % \toprule[0.9pt]
        \toprule
        \midrule        
            \multirow{2}{*}{\textbf{Methods}}&\multicolumn{3}{c}{\textbf{Dataset}} \\
              & \textbf{ACM}  & \textbf{DBLP} & \textbf{Freebase}  \\  
        \midrule
        
        GCN        & 70,496 &1,676,932&11,531,463\\
        Simple-HGN &1,264,838 &2,070,600& 12,177,806\\ 
        DGI     & 192 &256&448\\
        CPT-HG  & 192 &256&448\\
        GraphPrompt & 64 &64&64\\
        HGPrompt  & 67 &68&71\\
        \alg~  & 12 &24&58\\
        \midrule
        \bottomrule
        % \bottomrule[0.9pt]
        \end{tabular}
    }
\end{center}
\end{table}

\begin{table}[!t]
    \begin{center}
    \setlength{\tabcolsep}{10pt}
    \caption{Comparison of time efficiency per epoch for downstream node classification on the DBLP dataset.}
    \label{tab4}
    \scalebox{1.2}{
        \begin{tabular}{ c |  c  c  c }
        \toprule%[0.9pt]
        \midrule
            \multirow{2}{*}{\textbf{Methods}}&\multicolumn{2}{c}{\textbf{Times}} \\
              & \textbf{Train(s)}  & \textbf{Test(s)}  \\        
        \midrule
        GCN         & 0.0901 &0.1129\\
        Simple-HGN  & 0.0789 &0.0463\\ 
        GraphPrompt & 0.0568 &0.0513\\
        HGPrompt    & 0.0521 &0.0458\\
        \alg~       & 0.0460 &0.0339\\
        \midrule
        \bottomrule%[0.9pt]
        \end{tabular}
    }
    \end{center}
\end{table}

% \begin{figure}[!t]
%   \begin{minipage}[t]{0.49\linewidth}
%     \centering
%     \includegraphics[width=\linewidth]{Exp-Figs/ps_pf_our.pdf}
%     % \label{fig:image1}
%   \end{minipage}
%   \hfill
%   \begin{minipage}[t]{0.49\linewidth}
%     \centering
%     \includegraphics[width=\linewidth]{Exp-Figs/pure,best,random.pdf}
%     % \label{fig:image2}
%   \end{minipage}
%   \caption{Comparative analysis of model variants with component variations in disruption experimentation and Comparative Evaluation of Model Variants and Baseline Models.}
%   \label{fig:difAgg}
% \end{figure}

% \begin{figure}[!t]
%     \centering
%     \includegraphics[width=0.4\textwidth]{Exp-Figs/ps_pf_our.pdf} 
%     \caption{Comparative analysis of model variants with component variations in disruption experimentation.}
%     \label{fig:ps_pf}
% \end{figure}
% \begin{figure}[h]
%     \centering
%     \includegraphics[width=0.4\textwidth]{Exp-Figs/pure,best,random.pdf}
%     \caption{ Comparative Evaluation of Model Variants and Baseline Models.}
%     \label{fig:pure_best}
% \end{figure}

%% file: B-5-Experiments-RQ4.tex
\subsection{Hyperparameter Analyses (RQ4)}
\label{Sec: Hyperparameter Analyses} 

\begin{figure}[!t]
    \centering
    \includegraphics[width=0.4\textwidth]{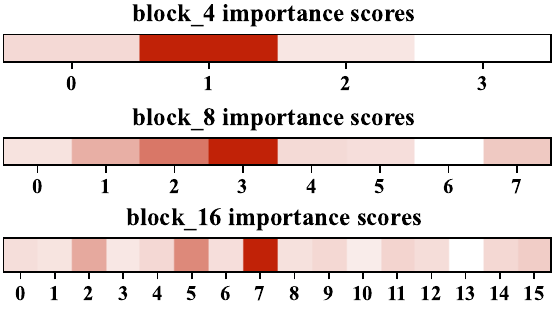}
    \caption{ Hotspot plot of importance scores for different numbers of blocks in feature prompts.}
    \label{fig:hotspot}
\end{figure}

We further validate the performance variation of \alg~at different shot numbers settings and the impact of the choice of the number of dividing blocks in the feature prompt $P_f$ on the importance score. 
The different division levels \textit{t} in Fig. \ref{fig:hotspot} affect the block scores of $P_f$ and thus the degree of pruning. 
For example, at \(t = 4\), the last block has a low score, which can lead to pruning. 
However, at \(t = 8\), the last two blocks overlap with the last block at \( t = 4\). 
Of these two blocks, only the sixth block has a low score, while the seventh block has a relatively high score. 
Thus, at \(t = 8\), only the sixth block is pruned, while the seventh block is retained. 
At \(t=16\), two small blocks fall below threshold, but score variance grows, making pruning less stable. 
A finer-grained division with a larger $t$ allows for more precise identification of inefficient features but may lead to unstable pruning due to increased score noise, while a coarser division with a smaller $t$ ensures smoother and more consistent pruning but risks removing useful information. 
Therefore, an appropriate t must be chosen to balance the granularity of the $P_f$ score without exceeding the limit of the $P_f$ dimension.

To answer \textbf{RQ4,} we can analyze the experimental results (Figs. \ref{fig:shot-graph} and \ref{fig:hotspot}) and find that \alg~exhibits different performances under different hyperparameter settings. 
Specifically, the advantage of our method is more obvious when the training sample size is small, which means that our method can obtain good results with the least amount of labeled data, which fits well with the prompt learning paradigm. 
Furthermore, we observe that the number of divided blocks in the feature prompt affects the importance score and hence the degree of pruning. 
These hyperparameter decisions directly affect the performance and efficacy of \alg, highlighting its sensitivity to hyperparameters. 
Therefore, when performing tasks with \alg, hyperparameters must be carefully tuned and optimized to ensure optimal performance and effectiveness.

% We further validate the performance variation of \alg~ at different shot nums settings and the impact of the choice of the number of dividing blocks in the feature prompt $P_f$ on the importance score. As shown in Fig. \ref{fig:shot-graph}, we find that the advantage of our method is more obvious when the number of training samples is small, which suggests that our method only needs a small amount of labeled data for training to achieve good results, which fits the problem solved by prompt learning.Therefore we chose  \(shot \:nums = 1\) as the experimental setup for the \textit{NC} task. Through Fig. \ref{fig:hotspot} we find that for $P_f$ , different number of divisions \(t\) will get different block evaluation results thus affecting the degree of pruning. For example, when \( t=4\), the score of the last block is too low and would typically be pruned. However, it can be observed that when \(t=8\), the last two blocks correspond to the same region as the last block at \(t=4\). Among these two blocks, only the score of the sixth block is low, while the score of the seventh block is relatively high. Therefore, at \(t=8\), only the sixth block would be pruned while the seventh block would be retained. We chose \(t = 16\) as our experimental setup because too small a k may result in not being able to evaluate the $P_f$ at a finer granularity, while too large a \(t\) may result in exceeding the dimension 64 of the $P_f$. 

%% file: B-6-Conclusion.tex
% ==================
% # IV. CONCLUSION #
% ==================
% \vspace{-0.1cm}
\section{Conclusion}
\label{sec:conclusion}

In this paper, we propose a novel graph prompting framework named \alg, which enhances the effectiveness of graph prompts and the richness of the information they convey. 
We adopt a graph prompt evaluation strategy aimed at assessing the importance of prompts for different tasks and demonstrate that different graph prompts and their respective components indeed vary in importance. 
By estimating prompt importance, we enhance the impact of graph prompts by pruning negative prompts and retuning positive prompts, thereby ensuring prompt efficiency and downstream task performance. 
Our proposed framework has been extensively tested on three public datasets to demonstrate its feasibility, effectiveness, and superiority.

%% file: A-main-IEEE.bbl
\begin{thebibliography}{10}
\providecommand{\url}[1]{#1}
\csname url@samestyle\endcsname
\providecommand{\newblock}{\relax}
\providecommand{\bibinfo}[2]{#2}
\providecommand{\BIBentrySTDinterwordspacing}{\spaceskip=0pt\relax}
\providecommand{\BIBentryALTinterwordstretchfactor}{4}
\providecommand{\BIBentryALTinterwordspacing}{\spaceskip=\fontdimen2\font plus
\BIBentryALTinterwordstretchfactor\fontdimen3\font minus \fontdimen4\font\relax}
\providecommand{\BIBforeignlanguage}[2]{{%
\expandafter\ifx\csname l@#1\endcsname\relax
\typeout{** WARNING: IEEEtran.bst: No hyphenation pattern has been}%
\typeout{** loaded for the language `#1'. Using the pattern for}%
\typeout{** the default language instead.}%
\else
\language=\csname l@#1\endcsname
\fi
#2}}
\providecommand{\BIBdecl}{\relax}
\BIBdecl

\bibitem{li2023survey}
X.~Li, L.~Sun, M.~Ling, and Y.~Peng, ``A survey of graph neural network based recommendation in social networks,'' \emph{Neurocomputing}, vol. 549, p. 126441, 2023.

\bibitem{xia2022multi}
L.~Xia, C.~Huang, Y.~Xu, P.~Dai, and L.~Bo, ``Multi-behavior graph neural networks for recommender system,'' \emph{in IEEE TNNLS}, vol.~35, no.~4, pp. 5473--5487, 2022.

\bibitem{reau2023deeprank}
M.~R{\'e}au, N.~Renaud, L.~C. Xue, and A.~M. Bonvin, ``Deeprank-gnn: a graph neural network framework to learn patterns in protein--protein interfaces,'' \emph{Bioinformatics}, vol.~39, no.~1, p. btac759, 2023.

\bibitem{he2022illuminati}
H.~He, Y.~Ji, and H.~H. Huang, ``Illuminati: Towards explaining graph neural networks for cybersecurity analysis,'' in \emph{IEEE EuroS\&P}, 2022, pp. 74--89.

\bibitem{wu2020comprehensive}
Z.~Wu, S.~Pan, F.~Chen, G.~Long, C.~Zhang, and S.~Y. Philip, ``A comprehensive survey on graph neural networks,'' \emph{in IEEE TNNLS}, vol.~32, no.~1, pp. 4--24, 2020.

\bibitem{chen2020smoothing}
J.~Chen, X.~Lin, H.~Xiong, Y.~Wu, H.~Zheng, and Q.~Xuan, ``Smoothing adversarial training for gnn,'' \emph{in IEEE TCSS}, 2020.

\bibitem{hu2023cost}
X.~Hu, H.~Chen, H.~Chen, S.~Liu, X.~Li, S.~Zhang, Y.~Wang, and X.~Xue, ``Cost-sensitive gnn-based imbalanced learning for mobile social network fraud detection,'' \emph{in IEEE TCSS}, 2023.

\bibitem{he2023label}
Y.~He, Y.~Zhang, F.~Yang, D.~Yan, and V.~S. Sheng, ``Label-dependent graph neural network,'' \emph{in IEEE TCSS}, 2023.

\bibitem{cheng2023wiener}
J.~Cheng, M.~Li, J.~Li, and F.~Tsung, ``Wiener graph deconvolutional network improves graph self-supervised learning,'' in \emph{AAAI}, vol.~37, no.~6, 2023, pp. 7131--7139.

\bibitem{jiang2021pre}
X.~Jiang, T.~Jia, Y.~Fang, C.~Shi, Z.~Lin, and H.~Wang, ``Pre-training on large-scale heterogeneous graph,'' in \emph{SIGKDD}, 2021, pp. 756--766.

\bibitem{jin2021node}
W.~Jin, T.~Derr, Y.~Wang, Y.~Ma, Z.~Liu, and J.~Tang, ``Node similarity preserving graph convolutional networks,'' in \emph{WSDM}, 2021, pp. 148--156.

\bibitem{chen2020generative}
M.~Chen, A.~Radford, R.~Child, J.~Wu, H.~Jun, D.~Luan, and I.~Sutskever, ``Generative pretraining from pixels,'' in \emph{ICML}, 2020, pp. 1691--1703.

\bibitem{dong2019unified}
L.~Dong, N.~Yang, W.~Wang, F.~Wei, X.~Liu, Y.~Wang, J.~Gao, M.~Zhou, and H.-W. Hon, ``Unified language model pre-training for natural language understanding and generation,'' \emph{in NeurIPS}, vol.~32, 2019.

\bibitem{brown2020language}
T.~Brown, B.~Mann, N.~Ryder, M.~Subbiah, J.~D. Kaplan, P.~Dhariwal, A.~Neelakantan, P.~Shyam, G.~Sastry, A.~Askell \emph{et~al.}, ``Language models are few-shot learners,'' \emph{in NeurIPS}, vol.~33, pp. 1877--1901, 2020.

\bibitem{liu2021p}
X.~Liu, K.~Ji, Y.~Fu, W.~L. Tam, Z.~Du, Z.~Yang, and J.~Tang, ``P-tuning v2: Prompt tuning can be comparable to fine-tuning universally across scales and tasks,'' \emph{arXiv preprint arXiv:2110.07602}, 2021.

\bibitem{sun2022gppt}
M.~Sun, K.~Zhou, X.~He, Y.~Wang, and X.~Wang, ``Gppt: Graph pre-training and prompt tuning to generalize graph neural networks,'' in \emph{SIGKDD}, 2022, pp. 1717--1727.

\bibitem{liu2023graphprompt}
Z.~Liu, X.~Yu, Y.~Fang, and X.~Zhang, ``Graphprompt: Unifying pre-training and downstream tasks for graph neural networks,'' in \emph{WWW}, 2023, pp. 417--428.

\bibitem{ge2023domain}
C.~Ge, R.~Huang, M.~Xie, Z.~Lai, S.~Song, S.~Li, and G.~Huang, ``Domain adaptation via prompt learning,'' \emph{in IEEE TNNLS}, 2023.

\bibitem{yu2024hgprompt}
X.~Yu, Y.~Fang, Z.~Liu, and X.~Zhang, ``Hgprompt: Bridging homogeneous and heterogeneous graphs for few-shot prompt learning,'' in \emph{AAAI}, vol.~38, no.~15, 2024, pp. 16\,578--16\,586.

\bibitem{ma2023hetgpt}
Y.~Ma, N.~Yan, J.~Li, M.~Mortazavi, and N.~V. Chawla, ``Hetgpt: Harnessing the power of prompt tuning in pre-trained heterogeneous graph neural networks,'' \emph{arXiv preprint arXiv:2310.15318}, 2023.

\bibitem{lv2025graphprompter}
R.~Lv, Z.~Zhang, K.~Zhang, Q.~Liu, W.~Gao, J.~Liu, J.~Yan, L.~Yue, and F.~Yao, ``Graphprompter: Multi-stage adaptive prompt optimization for graph in-context learning,'' \emph{arXiv preprint arXiv:2505.02027}, 2025.

\bibitem{fang2022prompt}
T.~Fang, Y.~M. Zhang, Y.~Yang, and C.~Wang, ``Prompt tuning for graph neural networks,'' 2022.

\bibitem{tan2023virtual}
Z.~Tan, R.~Guo, K.~Ding, and H.~Liu, ``Virtual node tuning for few-shot node classification,'' in \emph{SIGKDD}, 2023, pp. 2177--2188.

\bibitem{mh2024lvm}
D.~MH~Nguyen, H.~Nguyen, N.~Diep, T.~N. Pham, T.~Cao, B.~Nguyen, P.~Swoboda, N.~Ho, S.~Albarqouni, P.~Xie \emph{et~al.}, ``Lvm-med: Learning large-scale self-supervised vision models for medical imaging via second-order graph matching,'' \emph{in NeurIPS}, vol.~36, 2024.

\bibitem{li2023scigraphqa}
S.~Li and N.~Tajbakhsh, ``Scigraphqa: A large-scale synthetic multi-turn question-answering dataset for scientific graphs,'' \emph{arXiv preprint arXiv:2308.03349}, 2023.

\bibitem{gao2024protein}
Z.~Gao, X.~Sun, Z.~Liu, Y.~Li, H.~Cheng, and J.~Li, ``Protein multimer structure prediction via {PPI}-guided prompt learning,'' in \emph{ICLR}, 2024.

\bibitem{ma-etal-2022-xprompt}
F.~Ma, C.~Zhang, L.~Ren, J.~Wang, Q.~Wang, W.~Wu, X.~Quan, and D.~Song, ``{XP}rompt: Exploring the extreme of prompt tuning,'' in \emph{EMNLP}, 2022, pp. 11\,033--11\,047.

\bibitem{liu2020towards}
Z.~Liu, W.~Zhang, Y.~Fang, X.~Zhang, and S.~C. Hoi, ``Towards locality-aware meta-learning of tail node embeddings on networks,'' in \emph{CIKM}, 2020, pp. 975--984.

\bibitem{fang2024universal}
T.~Fang, Y.~Zhang, Y.~Yang, C.~Wang, and L.~Chen, ``Universal prompt tuning for graph neural networks,'' \emph{in NeurIPS}, vol.~36, 2024.

\bibitem{michel2019sixteen}
P.~Michel, O.~Levy, and G.~Neubig, ``Are sixteen heads really better than one?'' \emph{in NeurIPS}, vol.~32, 2019.

\bibitem{shen2020network}
X.~Shen, Q.~Dai, S.~Mao, F.-l. Chung, and K.-S. Choi, ``Network together: Node classification via cross-network deep network embedding,'' \emph{in IEEE TNNLS}, vol.~32, no.~5, pp. 1935--1948, 2020.

\bibitem{wang2020neighborhood}
Z.~Wang, Y.~Lei, and W.~Li, ``Neighborhood attention networks with adversarial learning for link prediction,'' \emph{in IEEE TNNLS}, vol.~32, no.~8, pp. 3653--3663, 2020.

\bibitem{DBLP:conf/iclr/FrankleC19}
J.~Frankle and M.~Carbin, ``The lottery ticket hypothesis: Finding sparse, trainable neural networks,'' in \emph{ICLR}, 2019.

\bibitem{kipf2016semi}
T.~N. Kipf and M.~Welling, ``Semi-supervised classification with graph convolutional networks,'' \emph{arXiv preprint arXiv:1609.02907}, 2016.

\bibitem{velickovic2017graph}
P.~Velickovic, G.~Cucurull, A.~Casanova, A.~Romero, P.~Lio, Y.~Bengio \emph{et~al.}, ``Graph attention networks,'' \emph{stat}, vol. 1050, no.~20, pp. 10--48\,550, 2017.

\bibitem{you2020graph}
Y.~You, T.~Chen, Y.~Sui, T.~Chen, Z.~Wang, and Y.~Shen, ``Graph contrastive learning with augmentations,'' \emph{in NeurIPS}, vol.~33, pp. 5812--5823, 2020.

\bibitem{ju2022commonsense}
J.~Ju, D.~Yang, and J.~Liu, ``Commonsense knowledge base completion with relational graph attention network and pre-trained language model,'' in \emph{CIKM}, 2022, pp. 4104--4108.

\bibitem{li2024graph}
X.~Li, Z.~Fan, F.~Huang, X.~Hu, Y.~Deng, L.~Wang, and X.~Zhao, ``Graph neural network with curriculum learning for imbalanced node classification,'' \emph{Neurocomputing}, vol. 574, p. 127229, 2024.

\bibitem{wei2021pooling}
L.~Wei, H.~Zhao, Q.~Yao, and Z.~He, ``Pooling architecture search for graph classification,'' in \emph{CIKM}, 2021, pp. 2091--2100.

\bibitem{tan2023bring}
Q.~Tan, X.~Zhang, N.~Liu, D.~Zha, L.~Li, R.~Chen, S.-H. Choi, and X.~Hu, ``Bring your own view: Graph neural networks for link prediction with personalized subgraph selection,'' in \emph{WSDM}, 2023, pp. 625--633.

\bibitem{devlin2018bert}
J.~Devlin, M.-W. Chang, K.~Lee, and K.~Toutanova, ``Bert: Pre-training of deep bidirectional transformers for language understanding,'' \emph{arXiv preprint arXiv:1810.04805}, 2018.

\bibitem{liu2021learning}
Z.~Liu, Y.~Shen, X.~Cheng, Q.~Li, J.~Wei, Z.~Zhang, D.~Wang, X.~Zeng, J.~Gu, and J.~Zhou, ``Learning representations of inactive users: A cross domain approach with graph neural networks,'' in \emph{CIKM}, 2021, pp. 3278--3282.

\bibitem{chen2022cross}
Y.~Chen, Y.~Zheng, Z.~Xu, T.~Tang, Z.~Tang, J.~Chen, and Y.~Liu, ``Cross-domain few-shot classification based on lightweight res2net and flexible gnn,'' \emph{Knowledge-based systems}, vol. 247, p. 108623, 2022.

\bibitem{sun2022does}
R.~Sun, H.~Dai, and A.~W. Yu, ``Does gnn pretraining help molecular representation?'' \emph{in NeurIPS}, vol.~35, pp. 12\,096--12\,109, 2022.

\bibitem{zhang2022robust}
Y.~Zhang, H.~Gao, J.~Pei, and H.~Huang, ``Robust self-supervised structural graph neural network for social network prediction,'' in \emph{WWW}, 2022, pp. 1352--1361.

\bibitem{sun2023all}
X.~Sun, H.~Cheng, J.~Li, B.~Liu, and J.~Guan, ``All in one: Multi-task prompting for graph neural networks,'' in \emph{SIGKDD}, 2023, pp. 2120--2131.

\bibitem{huang2024prodigy}
Q.~Huang, H.~Ren, P.~Chen, G.~Kr{\v{z}}manc, D.~Zeng, P.~S. Liang, and J.~Leskovec, ``Prodigy: Enabling in-context learning over graphs,'' \emph{in NeurIPS}, vol.~36, 2024.

\bibitem{chen2023ultra}
M.~Chen, Z.~Liu, C.~Liu, J.~Li, Q.~Mao, and J.~Sun, ``Ultra-dp: Unifying graph pre-training with multi-task graph dual prompt,'' \emph{arXiv preprint arXiv:2310.14845}, 2023.

\bibitem{wang2019heterogeneous}
X.~Wang, H.~Ji, C.~Shi, B.~Wang, Y.~Ye, P.~Cui, and P.~S. Yu, ``Heterogeneous graph attention network,'' in \emph{WWW}, 2019, pp. 2022--2032.

\bibitem{lv2021we}
Q.~Lv, M.~Ding, Q.~Liu, Y.~Chen, W.~Feng, S.~He, C.~Zhou, J.~Jiang, Y.~Dong, and J.~Tang, ``Are we really making much progress? revisiting, benchmarking and refining heterogeneous graph neural networks,'' in \emph{SIGKDD}, 2021, pp. 1150--1160.

\bibitem{bollacker2008freebase}
K.~Bollacker, C.~Evans, P.~Paritosh, T.~Sturge, and J.~Taylor, ``Freebase: a collaboratively created graph database for structuring human knowledge,'' in \emph{SIGMOD}, 2008, pp. 1247--1250.

\bibitem{velivckovic2018deep}
P.~Veli{\v{c}}kovi{\'c}, W.~Fedus, W.~L. Hamilton, P.~Li{\`o}, Y.~Bengio, and R.~D. Hjelm, ``Deep graph infomax,'' \emph{arXiv preprint arXiv:1809.10341}, 2018.

\bibitem{you2021graph}
Y.~You, T.~Chen, Y.~Shen, and Z.~Wang, ``Graph contrastive learning automated,'' in \emph{ICML}, 2021, pp. 12\,121--12\,132.

\bibitem{jiang2021contrastive}
X.~Jiang, Y.~Lu, Y.~Fang, and C.~Shi, ``Contrastive pre-training of gnns on heterogeneous graphs,'' in \emph{CIKM}, 2021, pp. 803--812.

\bibitem{wang2021self}
X.~Wang, N.~Liu, H.~Han, and C.~Shi, ``Self-supervised heterogeneous graph neural network with co-contrastive learning,'' in \emph{SIGKDD}, 2021, pp. 1726--1736.

\bibitem{wang2020graph}
N.~Wang, M.~Luo, K.~Ding, L.~Zhang, J.~Li, and Q.~Zheng, ``Graph few-shot learning with attribute matching,'' in \emph{CIKM}, 2020, pp. 1545--1554.

\bibitem{sun2023graph}
X.~Sun, J.~Zhang, X.~Wu, H.~Cheng, Y.~Xiong, and J.~Li, ``Graph prompt learning: A comprehensive survey and beyond,'' \emph{arXiv preprint arXiv:2311.16534}, 2023.

\bibitem{liu2021relative}
Z.~Liu, Y.~Fang, C.~Liu, and S.~C. Hoi, ``Relative and absolute location embedding for few-shot node classification on graph,'' in \emph{AAAI}, vol.~35, no.~5, 2021, pp. 4267--4275.

\end{thebibliography}
